\documentclass[sigconf, nonacm]{acmart}

\usepackage[T1]{fontenc}

\usepackage{subcaption}

\usepackage{tikz}
\usepackage{amsmath}
\usepackage{siunitx}
\usepackage{booktabs}
\usepackage{multirow}
\usepackage{enumitem}
\usepackage{mathtools}
\usepackage{balance}
\usepackage{pifont}

\usepackage{breakurl}           
\usepackage{xcolor}             
\hypersetup{                    
  colorlinks,
  linkcolor={green!80!black},
  citecolor={red!70!black},
  urlcolor={blue!70!black}
} 
\usepackage{appendix}

\DeclarePairedDelimiter{\ceil}{\lceil}{\rceil}

\newcommand{\cmark}{\ding{51}}%
\newcommand{\xmark}{\ding{55}}%
\newcommand{\remove}[1]{\red{\sout{#1}}}

\copyrightyear{2021} 
\acmYear{2021} 
\setcopyright{acmcopyright}\acmConference[ACSAC '21]{Annual Computer Security Applications Conference}{December 6--10, 2021}{Virtual Event, USA}
\acmBooktitle{Annual Computer Security Applications Conference (ACSAC '21), December 6--10, 2021, Virtual Event, USA}
\acmPrice{15.00}
\acmDOI{10.1145/3485832.3488016}
\acmISBN{978-1-4503-8579-4/21/12}

\usepackage[absolute]{textpos}
\begin{document}

\begin{textblock}{18}(1.8,2)
\noindent\Large This paper was accepted for publication at the Annual Computer Security Applications Conference (ACSAC) 2021.
\end{textblock}

%
\title{They See Me Rollin': Inherent Vulnerability of the Rolling Shutter in CMOS Image Sensors}

\author{Sebastian Köhler}
\affiliation{%
   \institution{University of Oxford}
   \country{}}
\email{sebastian.kohler@cs.ox.ac.uk}

\author{Giulio Lovisotto}
\affiliation{%
   \institution{University of Oxford}
   \country{}}
\email{giulio.lovisotto@cs.ox.ac.uk}

\author{Simon Birnbach}
\affiliation{%
   \institution{University of Oxford}
   \country{}}
\email{simon.birnbach@cs.ox.ac.uk}

\author{Richard Baker}
\affiliation{%
   \institution{University of Oxford}
   \country{}}
\email{richard.baker@cs.ox.ac.uk}

\author{Ivan Martinovic}
\affiliation{%
   \institution{University of Oxford}
   \country{}}
\email{ivan.martinovic@cs.ox.ac.uk}

\begin{abstract}

In this paper, we describe how the electronic rolling shutter in CMOS image sensors can be exploited using a bright, modulated light source (e.g., an inexpensive, off-the-shelf laser), to inject fine-grained image disruptions. 
We demonstrate the attack on seven different CMOS cameras, ranging from cheap IoT to semi-professional surveillance cameras, to highlight the wide applicability of the rolling shutter attack.
We model the fundamental factors affecting a rolling shutter attack in an uncontrolled setting.
We then perform an exhaustive evaluation of the attack's effect on the task of object detection, investigating the effect of attack parameters.
We validate our model against empirical data collected on two separate cameras, showing that by simply using information from the camera's datasheet the adversary can accurately predict the injected distortion size and optimize their attack accordingly.
We find that an adversary can hide up to 75\% of objects perceived by state-of-the-art detectors by selecting appropriate attack parameters. 
We also investigate the stealthiness of the attack in comparison to a na\"{i}ve camera blinding attack, showing that common image distortion metrics can not detect the attack presence.
Therefore, we present a new, accurate and lightweight enhancement to the backbone network of an object detector to recognize rolling shutter attacks.
Overall, our results indicate that rolling shutter attacks can substantially reduce the performance and reliability of vision-based intelligent systems.

\end{abstract}

\maketitle



\section{Introduction}

\begin{figure*}[t]
	\centering
	\begin{subfigure}[b]{.24\linewidth}
		\centering
		\includegraphics[width=.95\textwidth]{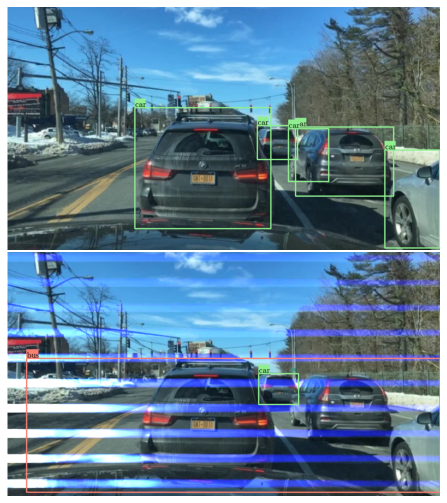}
		\caption{Object detection.}
		\label{fig:usecases-objectdetection} 
	\end{subfigure}%
	\begin{subfigure}[b]{.24\linewidth}
		\centering
		\includegraphics[width=.95\textwidth]{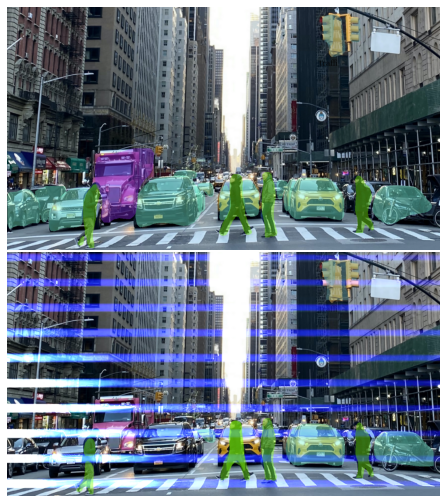}
		\caption{Instance segmentation.}
		\label{fig:usecases-segmentation} 
	\end{subfigure}%
	\begin{subfigure}[b]{.24\linewidth}
		\centering
		\includegraphics[width=.95\textwidth]{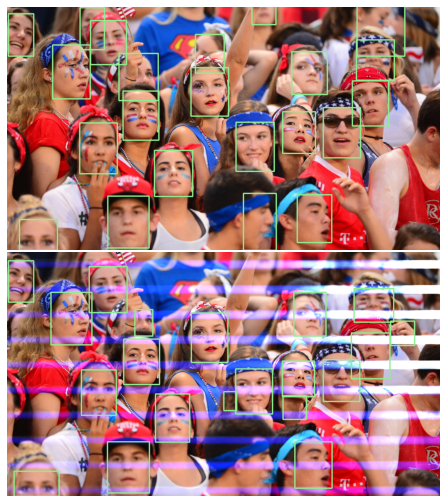}
		\caption{Face detection.}
		\label{fig:usecases-facedetection} 
	\end{subfigure}%
	\begin{subfigure}[b]{.24\linewidth}
		\centering
		\includegraphics[width=.95\textwidth]{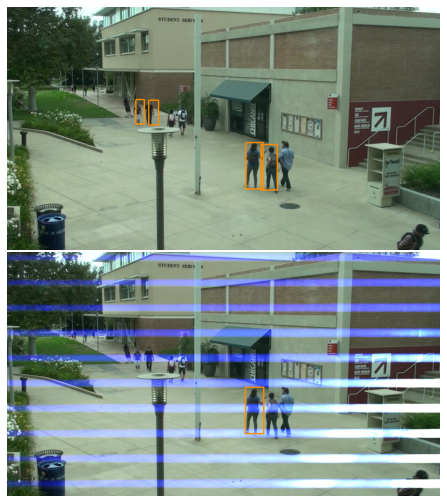}
		\caption{Person tracking.}
		\label{fig:usecases-persontracking} 
	\end{subfigure}%
	\caption{Computer vision use-cases and impact of rolling shutter attacks on predictions. The distortions introduced by the attack compromise the task's accuracy. From left to right, models are FasterRCNN~\cite{ren2015faster}, MaskRCNN~\cite{he2017mask}, RFCN~\cite{dai2016r} and SSD~\cite{liu2016ssd}.
	These distortions are more stealthy than simple blinding attacks, as demonstrated in Section~\ref{sec:comparison_with_blinding}.}
	\label{fig:usecases}
\end{figure*}

Digital cameras are undoubtedly ubiquitous today, with units now integrated into most mobile and IoT devices and already dominant in photography, monitoring, and surveillance.
With the ongoing integration of cameras into the decision-making process of intelligent systems, such as consumer home-security~\cite{nestcam,ring}, commercial security\cite{axis-website}, blanket surveillance~\cite{chinafacialrecog2019,londonfacialrecog2020}, construction site safety~\cite{everguardsentri}, retail~\cite{amazongo2020}, and vehicle automation~\cite{petit2015remote,yan2016can}, the integrity of the captured images is becoming a cornerstone of the correct behavior of the system.

Recent academic work has produced many examples of disrupting camera-based intelligent systems using projections on surfaces in the environment~\cite{nassi2019mobilbye,zhou2018invisible, lovisotto2020slap}, or the placement of printed patches~\cite{eykholt2018physical}.
Attacks inhibiting the use of camera footage are well known and repeatedly shown in different settings~\cite{petit2015remote,yan2016can}. 

Na\"{i}ve disruption of cameras using powerful light projected \textit{at the camera} is straightforward and often effective, shown in academic work~\cite{petit2015remote} and real-world settings~\cite{hkprotestorslasers2019}.
However, this disruption has previously been coarse-grained, only blinding the camera entirely.
While blinding attacks may cause significant disruption, the presence of blinding is easily detected as it generates substantial and sudden image changes.
Additionally, blinding attacks trigger significant exposure adjustments, where the camera has to reduce the time the sensor is exposed to compensate for the increasing amount of light, and are therefore generally considered suspicious~\cite{synology, blueiris}.

Yet, we show that extremely widespread image sensors can be exploited to obtain small-scale control over the captured images. This is because Complementary Metal-Oxide Semiconductor (CMOS) image sensors typically implement an electronic rolling shutter mechanism that captures the image in sequential rows instead of all-at-once. As CMOS image sensors provide a convenient balance between production costs and image quality, they have become the major sensor technology used in modern cameras. 

In this paper, we show how an adversary can exploit the electronic rolling shutter with a modulated light source and influence the images captured by the camera in a fine-grained way. 
The captured pixels' intensity can be varied arbitrarily in small bands of the image, down to a handful of pixel rows given sufficiently bright conditions. 
The attack is achievable with minimal preparation and no greater aiming requirements than a na\"{i}ve blinding attack, yet it can be further optimized based on knowledge of the target camera to enhance its effectiveness.
The attack is particularly effective when targeted at computer vision systems. 
As small-scale, high-frequency information plays a crucial role in these systems~\cite{wang2020high}, e.g., for edge detection, the injection of fine-grained distortion has a more profound effect on performance than na\"{i}ve blinding attacks. 
This is analogous to the injection of fine-grained, adversarial distortions into image data~\cite{goodfellow2014explaining}, or the physical world~\cite{song2018physical}. 

In this work, we introduce rolling shutter attacks and study them in practice.
We model the dynamics of the electronic rolling shutter, showing that all the parameters necessary to carry out an attack can typically be inferred from the camera's technical specifications.
We validate our modeling with empirical data collected from executing the rolling shutter attack on two cameras, showing how such data matches the expectations.
We then investigate the stealthiness of the rolling shutter attack, showing that compared to a blinding attack, it causes only small changes in a set of commonly-used metrics for measuring image distortion.
Nevertheless, we describe how rolling shutter artifacts have an outsize effect on the performance of a range of computer vision systems (illustrated in Figure~\ref{fig:usecases}). 
We evaluate the attack for the object detection task: first, we physically reproduce the attack on seven different cameras, followed by a broader evaluation showing that the rolling shutter attack significantly degrades the performance of state-of-the-art object detection networks, as shown in Figure~\ref{fig:usecases-objectdetection}.
Finally, we discuss countermeasures for the attack detection, including introducing a new countermeasure that re-uses part of the deep network for detection, and outline the attack limitations.
In summary, we make the following contributions:

\begin{itemize} \itemsep0pt
    \item We model the mechanisms behind the rolling shutter, we show that an attacker can accurately control the attack-injected distortions simply by inferring parameters from the target camera's technical specifications. We validate our modeling against empirical data by collecting a large dataset of attack executions with real hardware.
    \item We perform an extensive evaluation of the rolling shutter attack's effect on the object detection task, by partially simulating it on the BDD100K and the VIRAT dataset, showing that the integrity of object detection can be compromised.
    \item We assess the stealthiness of the attack across several metrics on the BDD100K video dataset, our analysis highlights how the rolling shutter attack is harder to detect in comparison to a blinding attack.
    \item We discuss a range of countermeasures and propose an accurate and lightweight enhancement to the backbone network of an object detector to detect rolling shutter attacks.
\end{itemize}


\section{Related Work}

Many cyber-physical systems rely on the physical measurements of a wide variety of sensors ranging from temperature sensors through cameras to Light Detection and Ranging (LiDAR) \cite{knapp2014industrial, chen-2013}.
Since the integrity of sensor measurements is crucial for systems to behave as intended,  research has been conducted to evaluate the vulnerabilities of sensors against signal injections on the physical layer.
Physical-layer attacks can be categorized by modality~\cite{giechaskiel2019sok} and substantial attention has been paid to each: electromagnetic radiation~\cite{tu2019trick,yan2016can}, acoustic waves~\cite{trippel2017walnut,zhang2017dolphinattack,xu2018analyzing} and optical emission~\cite{park2016ain,Sugawara2019,petit2015remote,Cao2019,shin2017illusion}.
In this work, we focus on the latter category. 

Due to their wide range, the focused brightness, and the targeting accuracy, lasers are often the first choice for attacks against optical sensors.
\cite{park2016ain} showed that an infrared laser could be used to manipulate the measurements of a drop sensor in a medical infusion pump.
In~\cite{Sugawara2019}, an amplitude-modulated laser was leveraged to inject voice commands into the microphone of a voice-controlled assistant.
Lasers have also been used to inject spoofed points into the sensor of a LiDAR system, leading to false-positive object detection~\cite{petit2015remote, Cao2019, shin2017illusion, sun2020towards}, and to na\"{i}vely blind cameras necessary for various Advanced Driver Assistance Systems (ADASs)~\cite{petit2015remote, yan2016can}.

In addition to attacks that directly target sensors on the physical layer, intensive research has been conducted that highlights the vulnerability of environment perception systems of modern cars. 
\cite{nassi2020phantom} demonstrated that cameras recognize objects projected onto the road as genuine objects or obstacles leading to unintended maneuvers.
Instead of projecting objects onto the road, \cite{man2020ghostimage} highlighted that it is possible to project objects directly onto the camera lens. 

Recently, \cite{li2020light} showed that the rolling shutter can be abused to create a backdoor in the training dataset for face recognition systems.
The closest work to our own is in~\cite{sayles2021invisible}, which demonstrated that the rolling shutter can be exploited to cause misclassifications in single-object scenes, in which the attacker controls the light source for an enclosed environment. 
We go beyond the findings in that work by relaxing some assumptions to a more generalized case, 
in which the attacker can not enjoy full control over the scene lighting and must contend with further uncontrolled environmental and camera specific parameters, such as the precise ambient light level, frame rate and exposure time, along with cluttered, dynamic environments. 
We analyze these factors in detail and develop a reliable model of required conditions for an effective real-world attack. 
We further validate our method on a range of cameras and show that it works consistently, beyond the single target evaluated in~\cite{sayles2021invisible}.
We study the problem of rolling shutter attacks in a general setting and describe their applicability and effectiveness in unconstrained conditions.

\section{Image Sensor Technology}

\subsection{Photodiode Arrays}\label{sec:sensor_part}

An image sensor is composed of a two-dimensional array of unit cells, also known as pixels. 
The central component of each pixel is a photodiode, which absorbs incident photons and converts them into a signal charge depending on the intensity of the light beam.
An image is captured by exposing each pixel in the photosensitive array to incident light for a certain amount of time.
This time is known as the integration period~\cite{durini-2019}, or \textit{exposure time}.
The longer the integration period, the more photons are converted into signal charge and the brighter the resulting image.
To capture an image and control its brightness, cameras implement an electronic shutter mechanism.
With electronic shutters, the image sensor is constantly exposed to light and signal charge is continuously generated.
However, the accumulated charge is flushed before reading, such that only the light gathered during the specified integration time, between the reset and readout scan, represents the final image~\cite{kinugasa1987}.
The length of the integration time is typically managed by an automatic exposure control system, in which the measurements from an integrated light meter are used to optimize the exposure period. 

\subsection{Scanning Element}\label{sec:scanning_element}

\begin{figure}[t]
	\centering
	\subcaptionbox*{} {
		\includegraphics[width=0.88\linewidth]{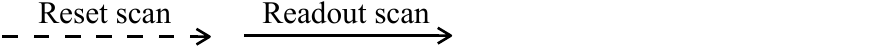}\vspace{-5mm}%
	} 
	\begin{subfigure}[b]{.44\linewidth}
		\centering
		\includegraphics[width=.9\textwidth]{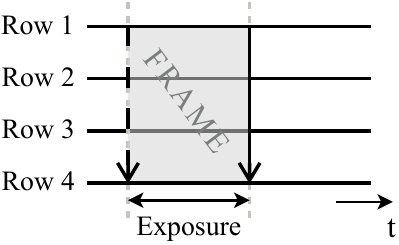}
		\caption{Electronic global shutter.}
		\label{fig:global_shutter_diagram} 
	\end{subfigure}%
	\quad
	\begin{subfigure}[b]{.44\linewidth}
		\centering
		\includegraphics[width=.9\textwidth]{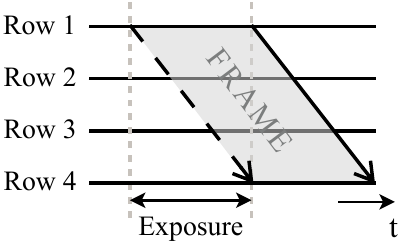}
		\caption{Electronic rolling shutter.}
		\label{fig:rolling_shutter_diagram} 
	\end{subfigure}%
	\caption{Visualization of the frame acquisition in the time-domain for the two shutter mechanisms.}
\end{figure}

The two most common image sensor technologies, Charge-Coupled Device (CCD) and Complementary Metal-Oxide Semiconductor (CMOS), use different techniques to recover the accumulated signal charge from the image array. 
The main difference is the arrangement of scanning elements and measurement units in the chip. 
Measurement units quantify the signal charge and convert it into electrical signals, whereas scanning elements control the order in which the captured signal charges are read~\cite{kuroda-2017}.
While CCD sensors have a single measurement unit, CMOS sensors have a measurement unit in each pixel~\cite{nakamura-2006}.
As a result, different electronic shutter implementations exist --- \textit{global} and \textit{rolling} electronic shutters.

\paragraph{Global Shutter}

Image sensors with an electronic global shutter simultaneously expose the entire photodiode array to incident light and also read out the accumulated signal charge from it at once.
In CCD image sensors the individual pixels are connected via vertical and horizontal shift registers that work as memory buffers.
The signal charge is temporarily stored inside the memory buffers, while the new integration period has already begun.
Figure \ref{fig:global_shutter_diagram} illustrates the readout of a CCD image sensor with an electronic global shutter in the time-domain. 
The use of such memory buffer weighs on production cost and chip size~\cite{Zhou1997, wany2003cmos}, which lead to CCD image sensors being most common in industrial and professional use~\cite{durini-2019}. 
In comparison, in CMOS sensors, the individual pixels are connected via metallic wires that cannot store the signal charge~\cite{kuroda-2017}.
Even though it is possible to implement an electronic global shutter in CMOS sensors by changing the circuit and adding additional memory buffer, it is often not economically viable and would defeat the main advantage --- cheap manufacturing that enables widespread use in consumer and semi-professional applications.

\paragraph{Rolling Shutter}

Image sensors with an electronic rolling shutter expose and read each row of the photodiode array one by one.
As a result, the readout of the last pixel row is slightly delayed compared to the first one.
An exemplary readout scan showing the time delay between the rows is depicted in Figure~\ref{fig:rolling_shutter_diagram}.
Since CMOS image sensors do not have a memory buffer, they can only implement an electronic rolling shutter and thus suffer from image distortions as a consequence of the delay between reading the first and last pixel row --- banding and geometric distortion~\cite{kuroda-2017}.
The most common phenomena are skewed or compressed objects, which can be observed particularly well when fast-moving objects are captured.
Another frequently occurring distortion is horizontal banding, which is especially common in indoor environments where dimmable LED lights are used.
Since LEDs are often dimmed by switching them on and off at a high frequency, bright horizontal bars might appear in the captured image~\cite{dietz2019shuttering}.
In our proposed rolling shutter attack, we exploit this phenomenon to inject controlled distortions into the captured image.

\section{Threat Model}\label{sec:threat_model}

In this paper, we focus on defining a general basis for a rolling shutter attack on camera-based intelligent systems.

\paragraph{Goal}

The adversary's general goal is to reduce the reliability and performance of camera-based intelligent systems in a way that is \textit{stealthier} than a blinding attack (denial-of-service).
In the simplest case, where the camera is within the attacker's reach, it is possible to use paint, bullets, or direct physical force to prevent a camera from working properly. 
However, these truly brute-force attacks are mitigated by keeping a camera out-of-reach or using tamper sensors in the housing. They are also unattractive as they risk the attacker being caught on camera while conducting the attack. 
While it is straightforward to blind a camera with a very bright light, this approach suffers from two key problems.
First, it is easily detectable with even simple methods such as `scene too bright/dark' thresholds or `image global change' comparisons, or by monitoring automatic exposure levels~\cite{bosch-2016-tamperdetection}. 
Second, in bright environments, it often provides incomplete disruption.
The image is completely disrupted in one area, but the effect rapidly diminishes away from that point; leaving much of the image unobscured. 
The residual color tint has little effect on computer vision systems such as object detection or face recognition, that typically make greater use of high-frequency image components~\cite{wang2020high}. 

By contrast, ours is a \textit{stealthy} attack that minimizes observable disruption to the system, while in fact impairing its function, as we demonstrate in more detail in Section~\ref{sec:comparison_with_blinding}.

\paragraph{Capabilities}

We assume the attacker has access only to off-the-shelf equipment to drive and modulate the laser and to aim it appropriately. 
We consider adversaries that have no control or access over the target camera, although they \textit{may} be able to obtain knowledge of the camera hardware specifications, either from a datasheet or from a replica device.
However, knowledge of camera parameters is not a requirement, as we explain in Section~\ref{sec:attack} and demonstrate on a range of cameras in Section~\ref{sec:exp-setup}.
Similar to previous work~\cite{petit2015remote, Cao2019, sun2020towards} and as shown in~\cite{caoautomated}, we consider that the adversary has line-of-sight with the target camera and the capabilities of aiming the laser at it.
In contrast to \cite{li2020light, sayles2021invisible}, which assume a rolling shutter attack in a static scene (i.e., single object, fixed distance and static lighting), we focus on a more general threat model.
Following the assumption that the adversary cannot access the video of the target camera, we consider the attack to be executed in a black-box setting without any feedback channel.
As such, distortions can only be injected at random locations, making it infeasible to target specific objects or apply optimization techniques.
However, our approach makes the attack applicable to dynamic multi-object scenes.

\section{Exploiting the Rolling Shutter}\label{sec:attack}

\begin{figure}[t]
	\centering
	\begin{subfigure}[b]{.30\linewidth}
		\centering
		\includegraphics[clip, trim=0cm 7.45cm 0cm 0cm,width=.85\textwidth]{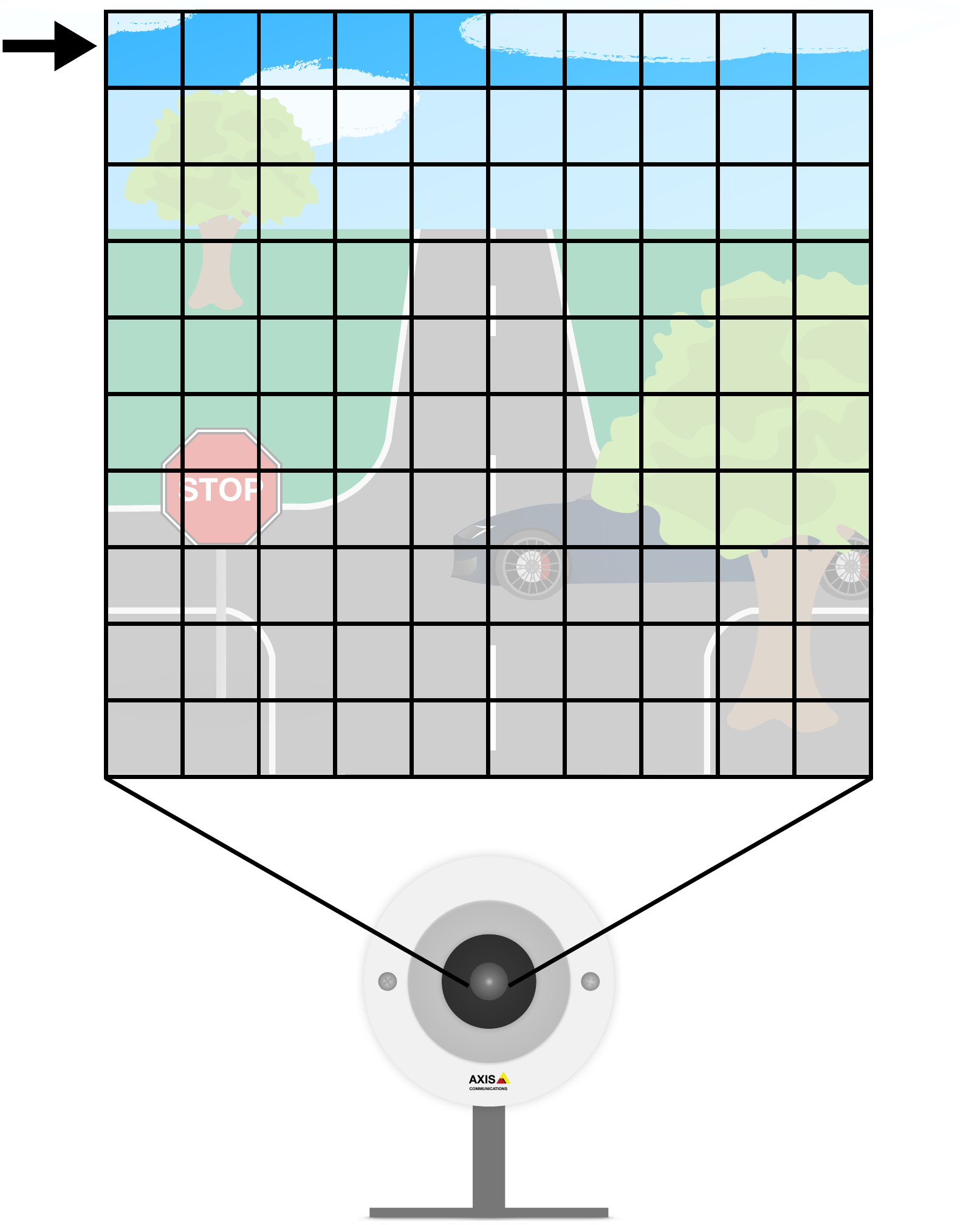}
		\caption{}
		\label{fig:attack-firstrow} 
	\end{subfigure}%
	\quad
	\begin{subfigure}[b]{.30\linewidth}
		\centering
		\includegraphics[clip, trim=0cm 7.45cm 0cm 0cm, width=.85\textwidth]{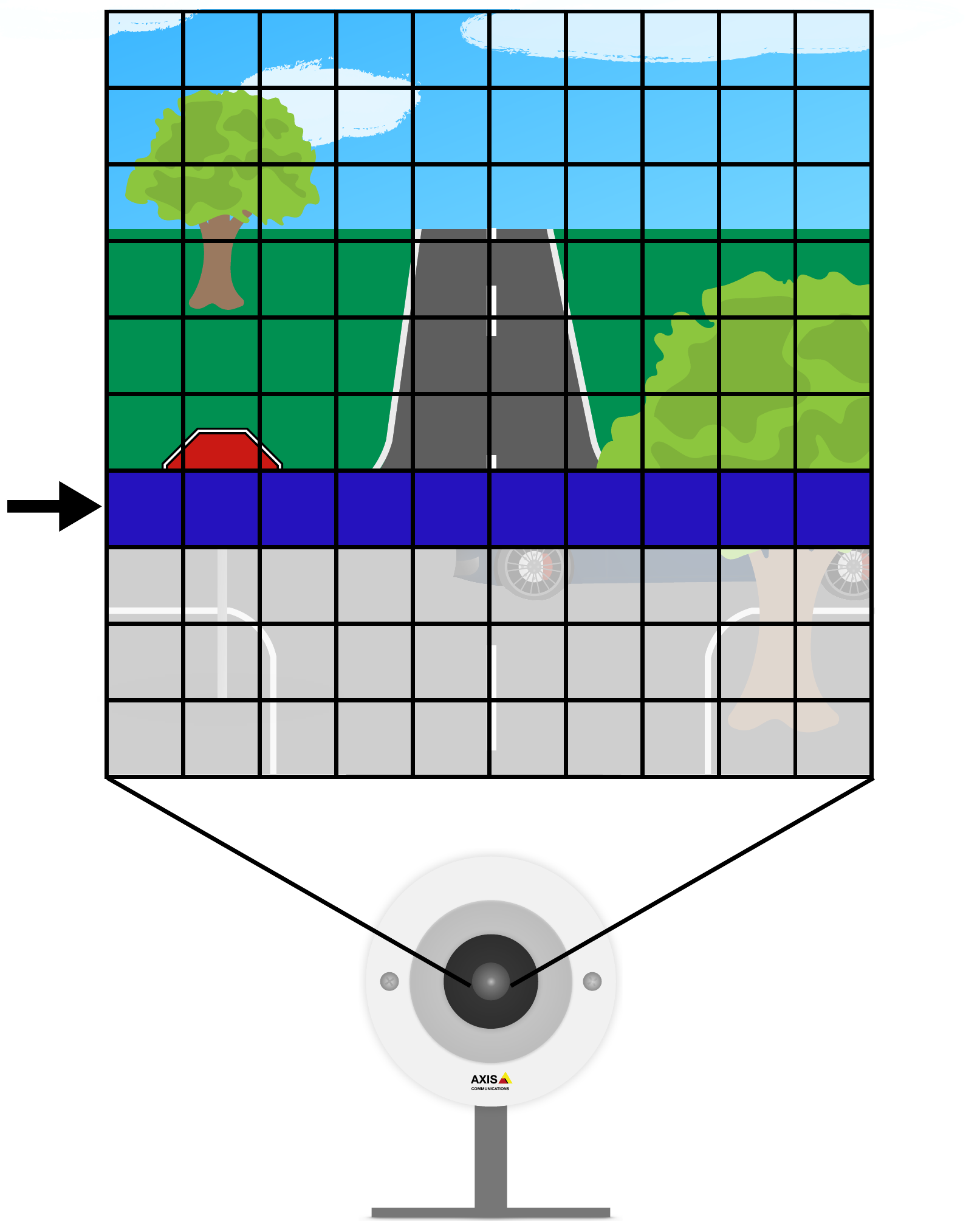}
		\caption{}
		\label{fig:attack-secondrow} 
	\end{subfigure}%
	\quad
	\begin{subfigure}[b]{.30\linewidth}
		\centering
		\includegraphics[clip, trim=0cm 7.45cm 0cm 0cm,width=.85\textwidth]{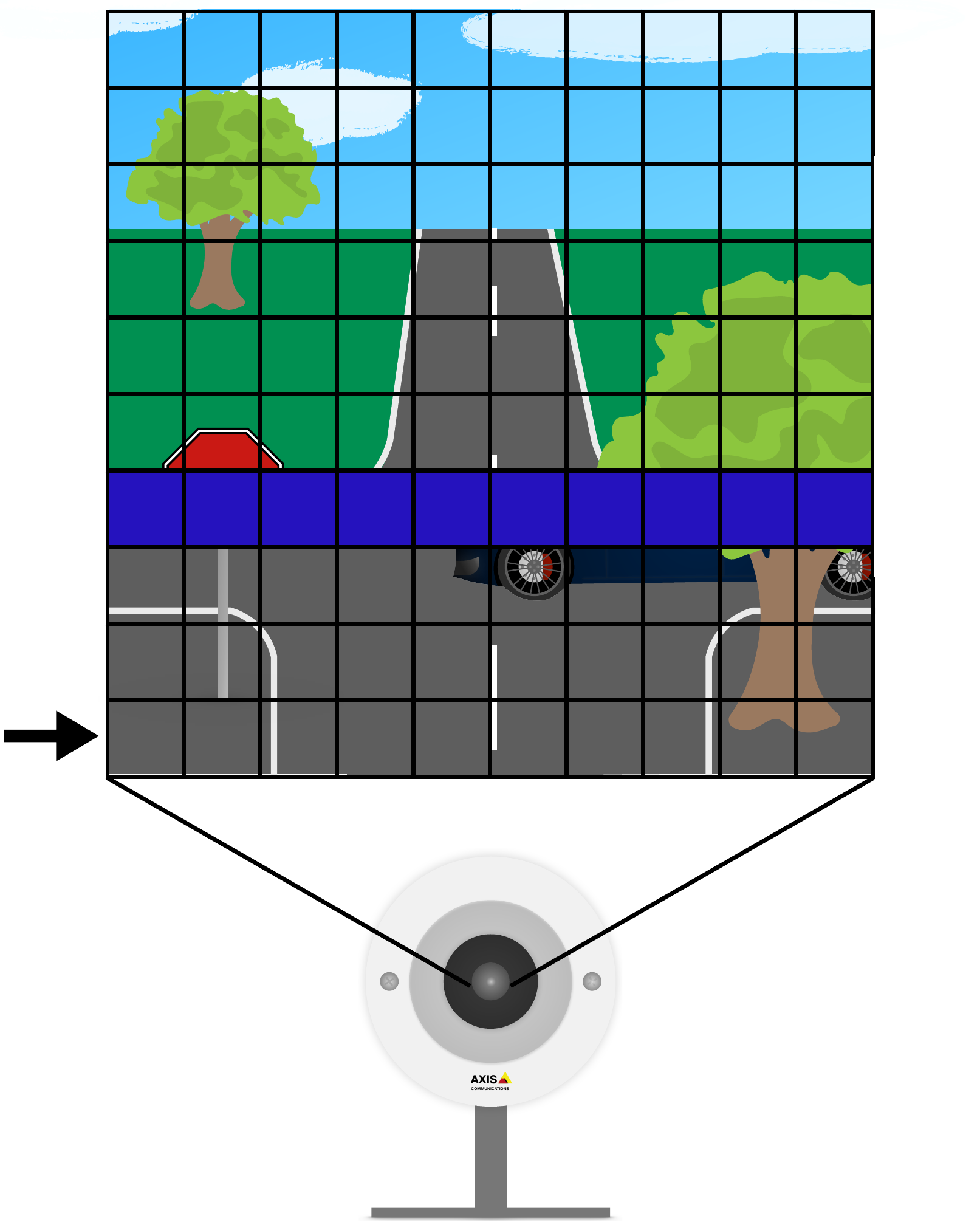}
		\caption{}
		\label{fig:attack-thirdrow} 
	\end{subfigure}%
	\caption{Illustration of the row-wise acquisition of a single frame. The arrow on the left indicates the row that is currently exposed and readout. While reading the frame, light is injected during the exposure of a row, leading to distortions in a small part of the image. }
	\label{fig:attack}
\end{figure}

\paragraph{Preliminaries} 
The rolling shutter attack is based on the consecutive exposure and readout of individual pixel rows. 
Each row is exposed for a short period $t_{exp}$, often in the range of microseconds, before the exposure of the next row starts. 
The time delay $\Delta{t}_{\text{rst}}$ between the reset pulses of consecutive lines can be exploited to inject noise that causes partial distortion of the captured image. 
If a modulated bright light, e.g., a laser, is pointed at the target camera, additional light can be selectively injected into rows of the image. 
The proportion of the image that is disturbed is controlled by the time $t_{on}$, which determines how long the laser is switched on (i.e., is emitting light). 
The attack is illustrated in Figure~\ref{fig:attack}, which depicts the row-wise acquisition of a single frame.
In Figure~\ref{fig:attack-firstrow}, the readout starts at the top-most row, the laser is inactive, and consequently, no image disturbances occur. 
At some point, the laser is turned on and the row that is currently exposed, e.g.,  the 7\textsuperscript{th} row in Figure~\ref{fig:attack-secondrow}, is illuminated and the original image information is perturbed. 
Before the next line is exposed, the laser is switched off, and the rest of the frame is captured without any further interference, see Figure~\ref{fig:attack-thirdrow}.

\subsection{Controlling the Distortion Size}\label{sec:controlling-the-distortion}

\begin{figure*}[t]
	\centering
	\begin{subfigure}[b]{.49\linewidth}
		\centering
		\includegraphics[width=.85\textwidth]{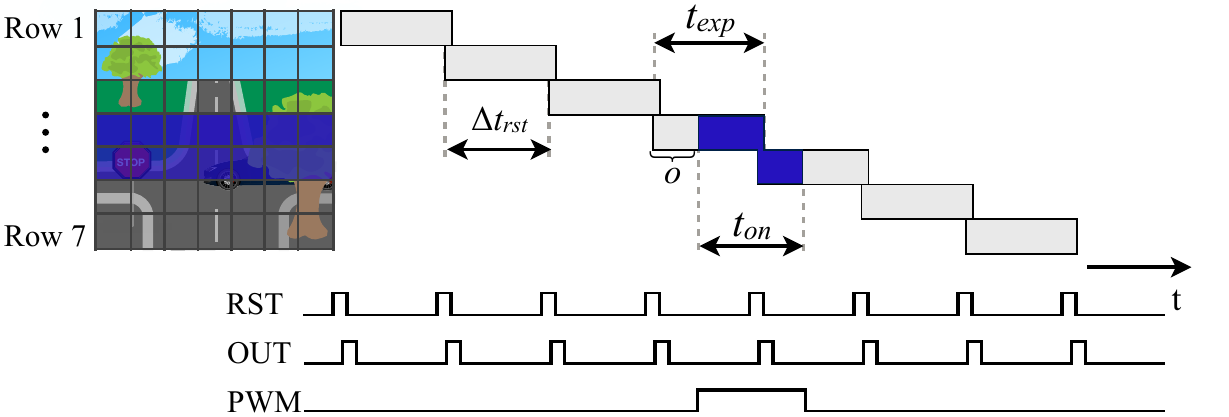}
		\caption{Short exposure time $t_{exp}$ = 32$\mu$s}
		\label{fig:overlap_attack_short_exp} 
	\end{subfigure}%
	\quad
	\begin{subfigure}[b]{.49\linewidth}
		\centering
		\includegraphics[width=.85\textwidth]{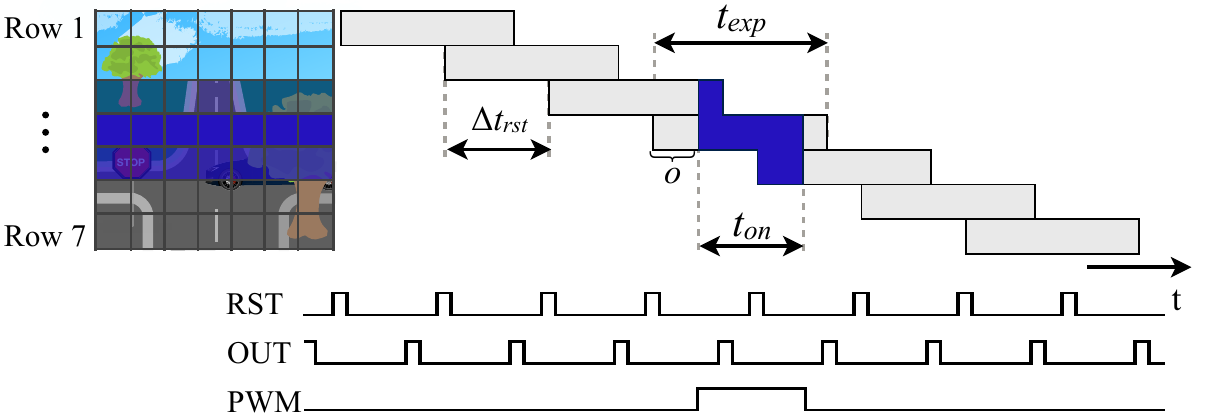}
		\caption{Long exposure time $t_{exp}$ = 50$\mu$s}
		\label{fig:overlap_attack_long_exp} 
	\end{subfigure}%
	\caption{Overview of the parameters that influence the size of the injected distortion. The signals RST and OUT represent the clock cycles for the reset and readout scan, the PWM signal indicates when the laser is on. 
	}
	\label{fig:attack_overlapping}
\end{figure*}

Shining a modulated light onto the image sensor will inject light in sets of adjacent pixel rows; we refer to each injection as a \textit{distortion}.
Depending on the use-case, the attacker might want to introduce large or narrow distortions. 
The appearance of the distortions depends on several factors.
Without loss of generality, we focus on rolling shutters that scroll top-to-bottom along the image sensor.
The same modeling applies for left-to-right rolling shutters, although the distortion would appear vertically rather than horizontally (as in~\cite{li2020light}).

\paragraph{Mechanisms behind Distortion Size}

The number of pixel rows that are illuminated by a distortion (\textit{distortion size)} depends on four factors: (i) the absolute exposure time of individual rows $t_{exp}$, (ii) the time the laser is switched on $t_{on}$, (iii) the time between $RST$ pulses $\Delta{t}_{\text{rst}}$, which is determined by the clock speed of the analog-to-digital converter, and (iv) the time delay between the last $RST$ pulse and the moment the laser is switched on, $o$.
Figure~\ref{fig:attack_overlapping} shows how the four parameters interact to define the distortion size.
Based on the four parameters, the adversary can precisely calculate the distortion size: 

\begin{equation} \label{eq:overlap_known_o}
N_{o} = \ceil[\bigg]{\frac{t_{exp}-o}{\Delta t_{rst}}} + \ceil[\bigg]{\frac{t_{on}+o}{\Delta t_{rst}}} - 1
\\
\text{, } o \in [0, \Delta t_{rst}).
\end{equation}

Since the attacked has no feedback channel, the laser cannot be synchronized with the rolling shutter of the target camera, so the offset $o$ is unknown. 
Consequentially, the attacker can only learn an upper and lower bound for the number of consecutive rows that will be illuminated, that is:

\begin{align}
N_{min} &= \ceil[\bigg]{\frac{t_{exp}}{\Delta t_{rst}}} + \ceil[\bigg]{\frac{t_{on}}{\Delta t_{rst}}} - 2 \\[1em]
\label{eq:overlap_min_max}
N_{max} &= \ceil[\bigg]{\frac{t_{exp}}{\Delta t_{rst}}} + \ceil[\bigg]{\frac{t_{on}}{\Delta t_{rst}}}, \;\;\; \text{with} \\[1em]
  &N_{min} \leq N_{o} \leq N_{max}. \nonumber
 \end{align}

In modern CMOS image sensors, the exposure time is generally much higher than the clock speed, $t_{exp}>\Delta t_{rst}$, meaning that the exposure times for consecutive rows overlap~\cite{qimaging}.
As a consequence, it is impossible to affect \textit{individual} pixel rows with the laser, as the additional illumination will affect multiple rows at once, depending on the current $t_{exp}$.
Figure~\ref{fig:attack_overlapping} shows the frame acquisition for two different $t_{exp}$, a short and a long one, showing how larger $t_{exp}$ leads to wider overlap across consecutive rows.

\paragraph{Distortion Repetition}

The adversary can increase the number of injected distortions per-frame by increasing the laser frequency $f$. 
Intuitively, if the laser frequency $f$ matches the frame rate $F$ of the camera, each camera frame will present exactly one injected distortion.
To cause interference at multiple parts in the frame at the same time, the attacker can increase $f$.
The number of distortions per frame can be calculated as follows: 

\begin{equation} \label{eq:number_of_bars}
 N_{D} = \frac{f}{F}.
\end{equation}

When $f$ is not an integer multiple of $F$, the location of the distortions will change over consecutive frames.
For example, when $f > F$, the distortions will scroll downwards across consecutive frames.
Depending on the attack scenario, the adversary can pick $f$ accordingly to enhance, reduce or even prevent the movement of the distortions.

\subsection{Estimating Parameters}\label{sec:estimating_parameters}
While the mechanism to control the distortion size is consistent, in practice each camera has different parameters.
This means that the adversary needs to estimate these parameters for a target camera to control the distortion size, using Equation~\ref{eq:overlap_min_max} and Equation~\ref{eq:number_of_bars}.
While the attack can be carried out without accurately estimating these parameters, a better estimate will lead to the attack outcome being more predictable (i.e., more effective for the attacker).
Across the introduced parameters, $t_{on}$ and $f$ are controlled directly by the adversary, $\Delta t_{rst}$ and $F$ are camera-dependent but static, $t_{exp}$ is camera-dependent and dynamic.
In the following, we describe how an adversary can infer the camera-dependent parameters using the information on the technical datasheet of a camera model.

\paragraph{Estimating $\Delta t_{rst}$ and $F$}
The operating frame rate $F$ is generally available in the camera specifications (common examples are 25fps or 30fps).
The clock speed $\Delta t_{rst}$ is also sometimes available, nevertheless, when it is not, it can be estimated precisely using $F$ and the full resolution of the image sensor:

\begin{equation} \label{eq:delta_t_rst}
 \Delta t_{rst} = \frac{1}{F \cdot N_{rows}},
\end{equation}

where $N_{rows}$ is the number of pixel rows of the image sensor. 
For some cameras, the size of the physical resolution of the photodiode array (total resolution) is larger than the maximum resolution used to record videos.
For example, a 1920$\times$1080 full-HD video may have been recorded on a camera with $N_{\text{rows}}>1080$.
The rolling shutter will still go through \textit{all} $N_{\text{rows}}$ pixel rows to capture additional image data, such as color information, but ignore a portion of them to read frames in a smaller resolution than the physical resolution; the ignored rows are known as \textit{dead areas}.
If the target image sensor has dead areas, Equation \ref{eq:number_of_bars}, used to calculate the number of injected distortions, has to be extended by the ratio of visible rows $N_{visible}$ and total rows $N_{rows}$:

\begin{equation} \label{eq:number_of_bars_dead_area}
 N_{D} = \frac{f}{F} \cdot \frac{N_{visible}}{N_{rows}}.
\end{equation}
The sensor's total resolution $N_{rows}$ and video resolution $N_{visible}$ can be found in the camera's technical specifications.

\paragraph{Estimating $t_{exp}$}
During operation, $t_{exp}$ is set automatically by an auto-exposure mechanism that balances the camera's $t_{exp}$ with the current ambient light level: dimmer scenes will require larger $t_{exp}$ to produce sufficient quality frames.
Cameras have differing light sensitivities and lens apertures, and therefore use different $t_{exp}$ for the same lighting conditions.
To estimate $t_{exp}$, the adversary can use the \textit{minimal luminous exposure} parameter $H_v$ required by the camera to capture a usable frame and measure the current ambient light level $E_v$, with a light meter, at the time of the attack.
Based on $E_v$ and $H_v$, the exposure time $t_{exp}$ can be calculated:
\begin{equation} \label{eq:estimate_texp}
 t_{exp} = \frac{H_v}{E_v}.
\end{equation}
In Appendix~\ref{sec:estimate_exp}, we show examples of how we accurately estimated $t_{exp}$. 
In case $H_v$ is not available, the adversary can measure the ambient light level $E_v$ versus $t_{exp}$ relationship by using a replica of the target camera and performing a set of measurements in controlled conditions.
At the time of the attack, the adversary needs to measure $E_v$, lookup the respective $t_{exp}$ that would be set by the target camera, and adjust the attack parameters accordingly.
In Section~\ref{sec:acc_of_shutter_m}, we demonstrate that imprecise estimates of $t_{exp}$ do not affect the distortion size substantially.

\section{Rolling Shutter Model Evaluation} \label{sec:rsmeval}

This section describes the setup we used to execute the attack on a variety of cameras, a deeper analysis on the factors affecting the attack and the validation of our model presented in Section~\ref{sec:attack}.

\subsection{Preliminaries}\label{sec:exp-setup}

\paragraph{Equipment}

\begin{figure}[t]
  \centering
  \includegraphics[width=0.8\linewidth]{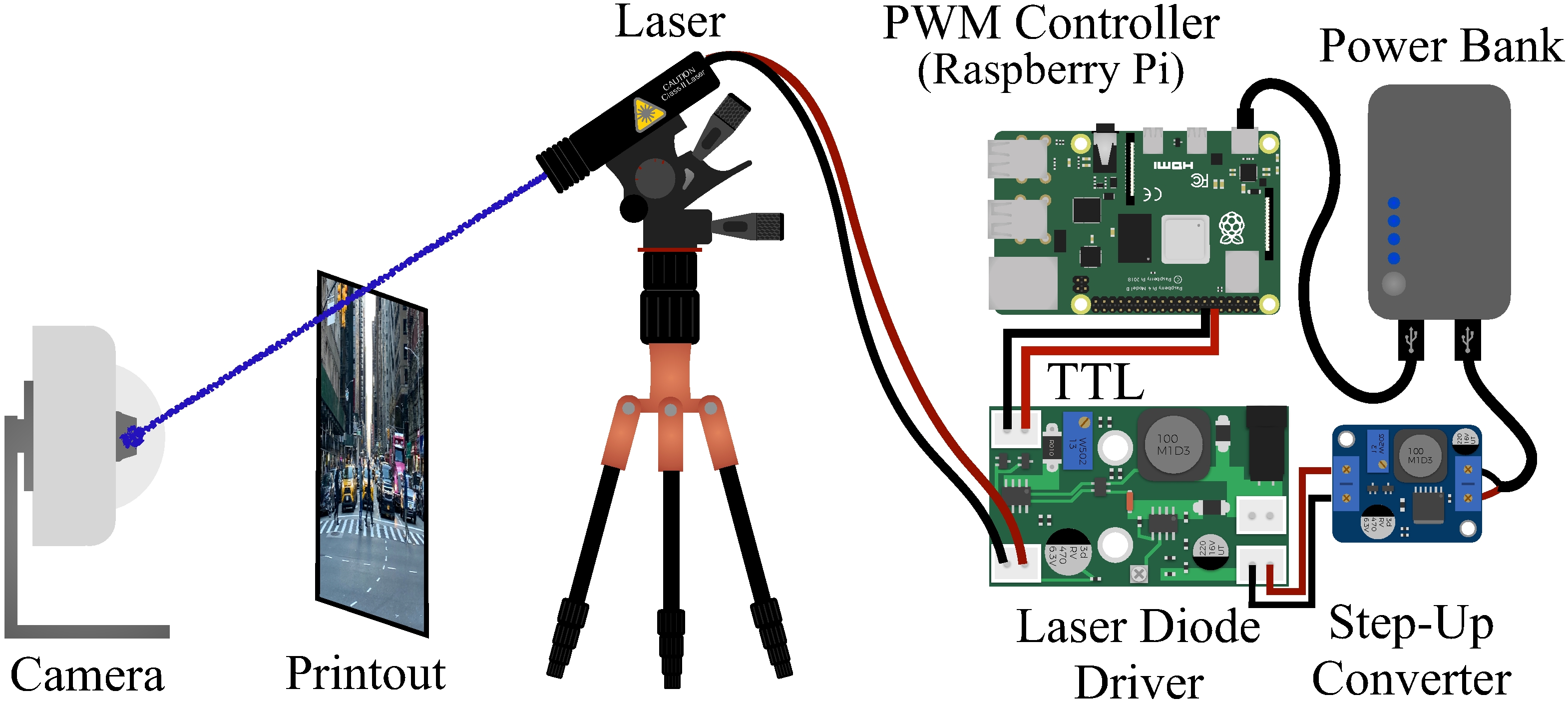}
  \caption{The experimental setup we used for the collection of the different patterns and the evaluation of the attack on various cameras.} 
  \label{fig:experimental_setup} 
\end{figure}

We only used off-the-shelf hardware that we bought online for less than \$90. 
An overview of the different components is depicted in Figure~\ref{fig:experimental_setup}.
We tested different lasers in terms of quality, wavelength (color), and power.
The lasers ranged from cheap 650nm (red), 535nm (green) and 405nm (violet/blue) laser pointers with an output power of 5mW~\cite{amazon-cheap-laser}, to more expensive and powerful semi-professional 532nm (green) and 445nm (blue) continuous wave (CW) diodes with power ratings of up to 1.8W~\cite{amazon-blue-laser}.
We noticed that an adversary might use green lasers to leverage the fact that most image sensors use a Bayer-Matrix color filter array and are therefore more sensitive to wavelengths at around 530nm~\cite{bayer1976colour}.

\paragraph{Laser Modulation} 

The laser was modulated using a laser driver with Transistor-Transistor-Logic that was connected to the hardware Pulse Width Modulation (PWM) controller of a Raspberry Pi.
A PWM modulated signal has two variables that can be changed --- the frequency and the duty cycle.
In our case, the frequency specifies how often the laser is switched on.
The duty cycle, given in percent, determines the duration of a pulse, i.e., how long the laser is switched on each time the signal is set to high. 
In other words, the duty cycle specifies $t_{on}$. 
Since $t_{on}$ depends on both frequency $f$ and the duty cycle, we will use $t_{on}$ in the rest of the paper rather than the duty cycle to make it easier to compare the injected distortions. 
Equation \ref{eq:on_time} shows how $t_{on}$ is calculated, given the frequency $f$ and the duty cycle $D$. 

\begin{equation} \label{eq:on_time}
t_{on} = \frac{1}{f}D.
\end{equation}

\paragraph{Applicability to Different Cameras}\label{sec:attack_transferability}

\begin{table}
	\centering
	\footnotesize
	\caption{Cameras we attempted to reproduce the attack on. The last column indicates whether we could reproduce the rolling shutter attack for the camera (\cmark) or not (\xmark).}
	\begin{tabular}{lcccc}
		\toprule
		\textbf{Camera} & \textbf{Resolution} & \textbf{$F$} & \textbf{Shutter} & \textbf{Success} \\\hline
		Logitech C922 & 1920$\times$1080 & 30 & rolling & \cmark\\
		Axis M3045-V & 1920$\times$1080 & 25 & rolling & \cmark\\
		YI Home Camera 1080p & 	1920$\times$1080 &24 & rolling & \cmark \\
		V380 Camera 720P& 1280$\times$720 &25	&  rolling & \cmark\\
		Google Nest & 1920$\times$1080 & 24  & rolling  & \cmark\\
		Netgear Arlo & 1280$\times$720 & 24  & rolling & \cmark\\
		Imaging Source DFM 25G445  & 1280$\times$960  & 25	&  global & \xmark\\
		Apple iPhone 7 & 3840$\times$2160 & 30 & rolling  & \cmark\\\bottomrule
	
	\end{tabular}
	\label{tab:tested_cameras}
\end{table}

We tested whether we could reproduce the rolling shutter attack on seven different cameras with CMOS image sensors, ranging from cheap IoT cameras (YI Home Camera) over mid-range smart cameras (Google Nest) to semi-professional surveillance (Axis M3045-V) and smartphone cameras (Apple iPhone 7).
For comparison, we also examined a camera with a CCD image sensor and an electronic global shutter mechanism.
We printed an image showing a busy intersection with multiple objects, such as pedestrians and vehicles, placed it in front of each camera and captured a frame during normal operation as well as while pointing a modulated laser at it.
The experimental setup is depicted in Figure~\ref{fig:experimental_setup} and examples of resulting attack frames are shown in Figure~\ref{fig:results_cameras}.
Table~\ref{tab:tested_cameras} provides an overview of all tested cameras and whether we were able to physically inject rolling shutter patterns.

\paragraph{In-depth Rolling Shutter Pattern Collection}\label{sec:experiment-protocol}

\begin{table}[t]
	\centering
	\footnotesize
	\caption{Attack parameters we collected for the Logitech and Axis cameras. The last column reports whether the parameter can be controlled by the adversary at attack time. We used a blue-colored laser with 1.8W of power.}
	\begin{tabular}{lccc}
		\toprule
		& \textbf{Range} & & \textbf{Adversary}\\ 
		\textbf{Param} & \textbf{[min, max]}  & \textbf{Symbol}&\textbf{Controlled} \\ \hline
		\textit{Camera Exposure} & & $t_{exp}$ & \xmark \\
		\hspace{.2cm} - Logitech & [100$\mu$s, 2,500$\mu$s] &  &  \\
		\hspace{.2cm} - Axis & [32$\mu$s, 1,000$\mu$s] &  &  \\
		\textit{Laser on-time} &  & $t_{on}$ & \cmark \\
		\hspace{.2cm} - Logitech & [320$\mu$s, 16,000$\mu$s] &  &  \\
		\hspace{.2cm} - Axis & [50$\mu$s, 400$\mu$s] &  &  \\
		\textit{Laser Frequency } & & $f$ &  \cmark   \\
		\hspace{.2cm} - Logitech & [30Hz, 900Hz] &  &  \\
		\hspace{.2cm} - Axis & [25Hz, 750Hz] &  &  \\
		\textit{Camera fps} & & $F$ &  \xmark   \\
		\hspace{.2cm} - Logitech & 30fps & &  \\
		\hspace{.2cm} - Axis & 25fps & &  \\ 
		\bottomrule
	\end{tabular}
	
	\label{tab:experiment_parameters}
\end{table}

To further analyze the factors affecting the attack, we picked two different cameras which allowed us to control the camera exposure time $t_{exp}$: the Logitech C922, a common webcam, and the Axis M3045-V, a dome surveillance camera.
We systematically reproduced the attack in controlled conditions by changing $t_{exp}$ and $t_{on}$, which affect the resulting distortions.
We placed the camera and the laser in a closed environment with little or no light, and collected a set of videos by varying the parameters of the attack, which are reported in Table~\ref{tab:experiment_parameters}.
We set the laser frequency $f$ to be slightly offset from the camera frame rate $F$ so that over consecutive video frames, the location of the pattern slowly iterated over the image rows.
This way, the set of collected frames covered all possible locations of the distortion in the image.
Using a mask that filtered out pixels with a value per color channel $\leq$ 10 (range [0, 255]), we extracted the generated \textit{pattern} of distortions from the captured video for each parameter configuration.

\subsection{Accuracy of the Shutter Model}\label{sec:acc_of_shutter_m}

\begin{figure}[t]
  \centering
  \includegraphics[width=.85\linewidth]{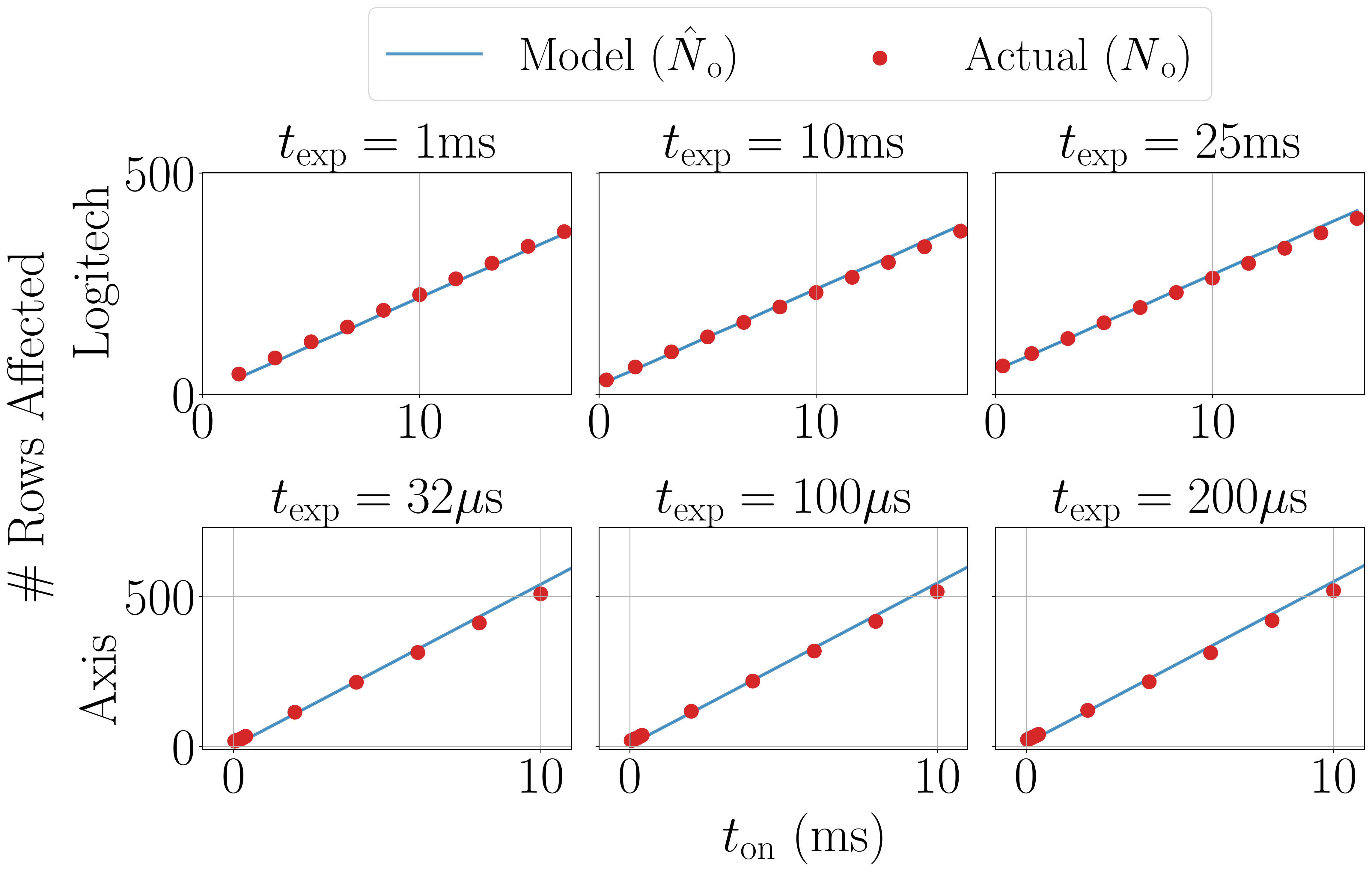}
  \caption{Comparison between the model-predicted number of affected rows and the empirical value found by measuring the number in practice on the two cameras.} 
  \label{fig:sec6-no-rows-affected} 
\end{figure}

Here, we show how the equations introduced in Section~\ref{sec:controlling-the-distortion} accurately model the number of rows affected by the distortion.

\paragraph{Estimated vs Empirical $N_o$} 
To estimate $N_o$ with Equation~\ref{eq:overlap_known_o} the adversary only needs $\Delta{t_{rst}}$; as $t_{on}$ is set by them and $t_{exp}$ is known in this experiment (in the next paragraph we analyze the estimation of $t_{exp}$).
We used Equation~\ref{eq:delta_t_rst} to estimate $\Delta{t_{rst}}$.
For the Logitech camera, this led to $\Delta{t_{rst}}=46.3\mu s$.
In contrast, the Axis camera has dead areas and requires $N_{rows}$ to be determined before $\Delta{t_{rst}}$ can be calculated.
For most cameras and image sensors, manufacturers provide detailed information about the number of total, effective and recording pixels.
However, for the Axis camera used in our experiment this information was not available.
By calculating the ratio between visible and invisible distortions, we estimated $N_{rows}$ to be 2,160 (1,080 visible and 1,080 invisible rows).
We therefore used $N_{rows}=2160$ for the Axis camera, resulting in $\Delta{t_{rst}}=18.5\mu s$.
The results in Figure~\ref{fig:sec6-no-rows-affected}, show how the prediction of our model closely matches the empirical outcome of the attack, for various $t_{exp}$, $t_{on}$, and the two considered cameras.
We find this model to be precise regardless of the part of the image that is being hit.
Figure~\ref{fig:sec6-no-rows-affected} also shows the confidence intervals for the number of rows affected by the attack, computed over every frame of the collected videos.

\paragraph{Incorrect Estimation of $t_{exp}$}

\begin{figure}[t]
  \centering
  \includegraphics[width=0.9\linewidth]{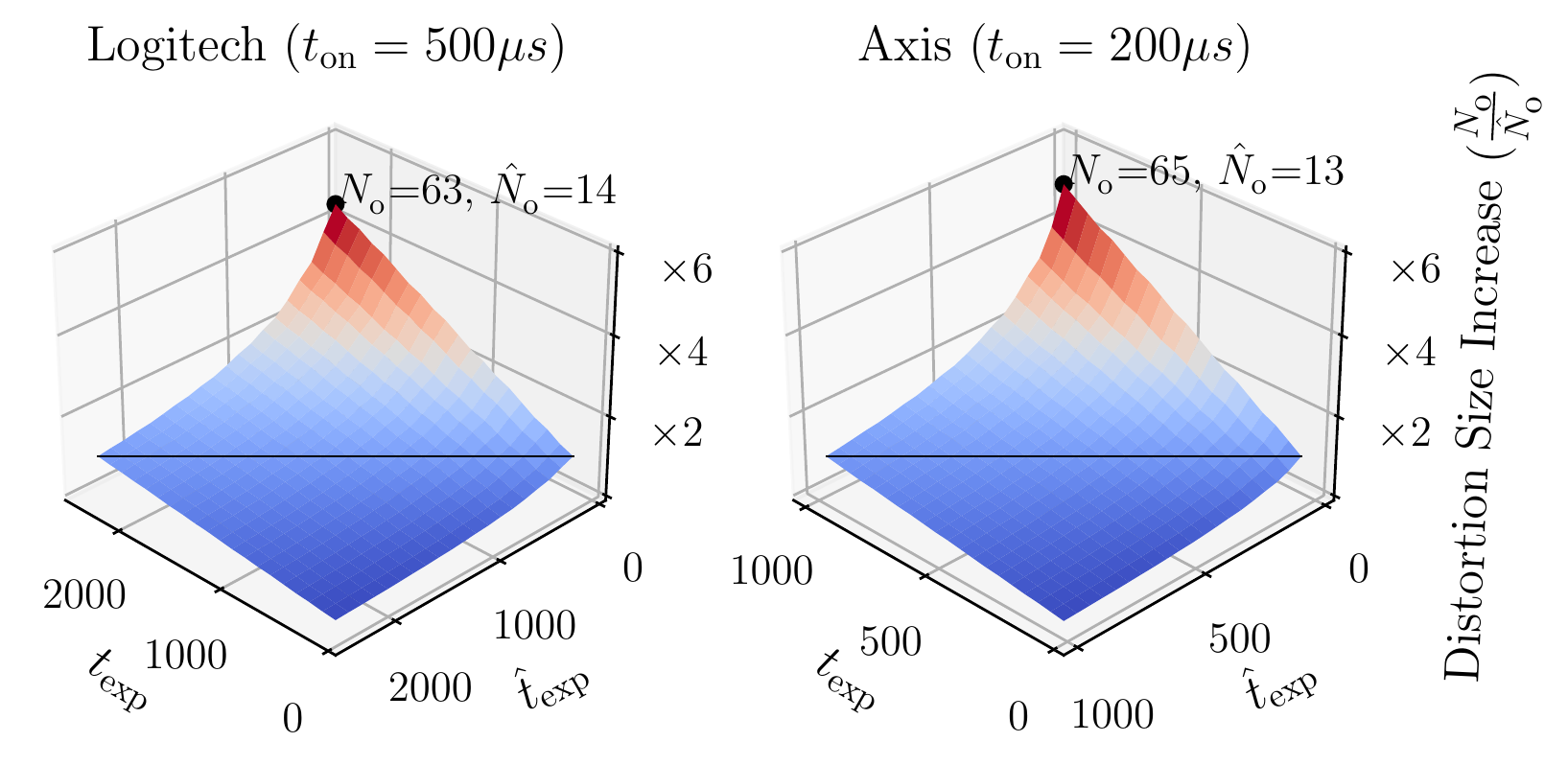}
  \caption{Increment in the expected distortion size $\hat{N}_{o}$ and actual size $N_{o}$ as the adversary's exposure time estimate $t_{exp}$ diverges from the true exposure time value $\hat{t}_{exp}$.} 
  \label{fig:sec6-incorrect-texp} 
\end{figure}

As mentioned in Section~\ref{sec:estimating_parameters}, the estimation of $t_{exp}$ can be imprecise if information from the datasheet is missing, leading to changes in the resulting distortion size $N_{o}$.
Using the rolling shutter model of the previous section, we computed the effect of an incorrect estimate of $t_{exp}$ on the resulting $N_{o}$.
We used $N_{max}$ (see Equation~\ref{eq:overlap_min_max}) as the effect of $o$ is negligible.
Given an estimate of $t_{exp}$ as $\hat{t}_{exp}$, and the resulting actual and estimated number of affected rows as $N_{o}$ and $\hat{N}_{o}$, we report the relationship between these four variables in Figure~\ref{fig:sec6-incorrect-texp}.
In the worst case, where the adversary underestimates $\hat{t}_{exp}$ and $t_{exp}$ is longest, this can lead to a distortion that is over 4$\times$ larger than expected.
The relationship depicted in Figure~\ref{fig:sec6-incorrect-texp} also dependents on $t_{on}$.
Additional plots for different $t_{on}$ are shown in Appendix~\ref{sec:estimate_texp_incorrect}.
Nevertheless, we find that roughly 70\% of the difference in distortion size is within a factor of two of the expected value (i.e., either twice as small or twice as large).
This shows that large deviations occur only when the adversary's estimate is far off the real exposure value, showing that the attack's outcome is quite predictable under an educated guess.

\section{Attack Evaluation}\label{sec:attacking-object-detection}

\begin{figure}[t]
	\centering
	\includegraphics[width=1\linewidth]{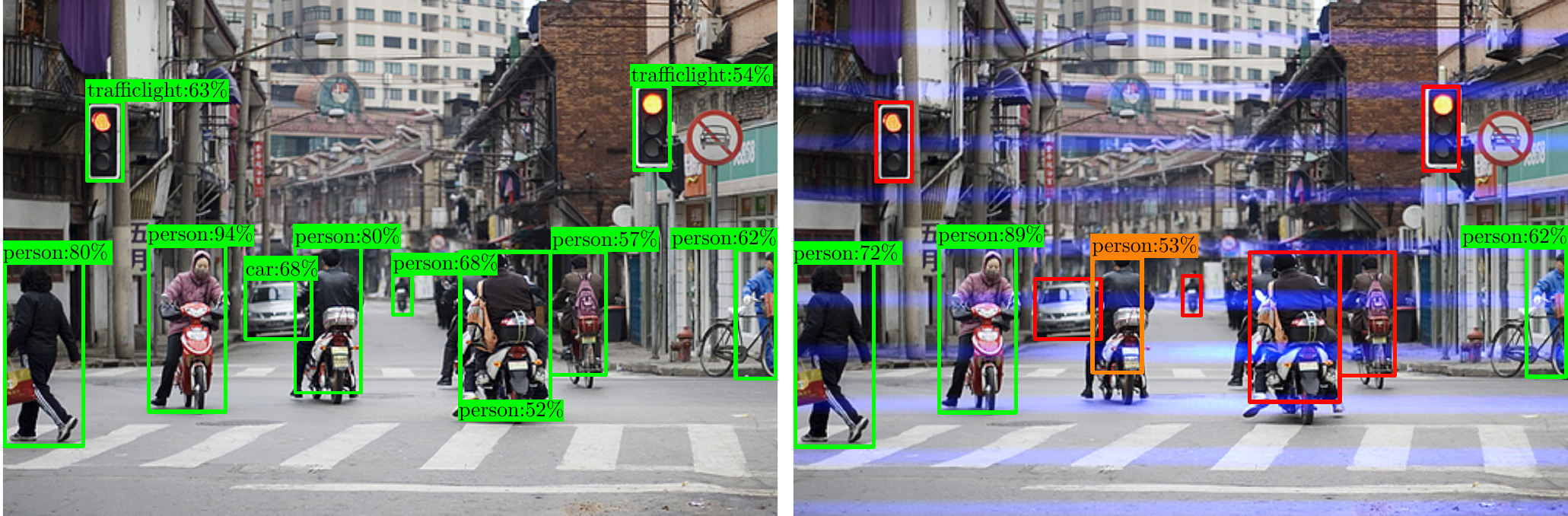}
	\caption{Results for SSD under normal operation (left) and with an attack pattern overlayed (right). The different colors indicate how an object is affected by the attack: red=hidden, orange=misplaced, green=unaltered.} 
	\label{fig:object_detection_vis}
	\vspace{-0.1cm} 
\end{figure}

This section presents the impact of the rolling shutter attack on the \textit{object detection} task, i.e., locating and identifying semantic objects.
In addition, it compares the amount of interference introduced by our attack with a blinding attack.
As described in our threat model in Section~\ref{sec:threat_model}, fine-grained optimizations are infeasible, thus the adversary chooses attack parameters by quantifying the attack success offline.

\subsection{Targeting Object Detection}\label{sec:evaluation_setup}

\paragraph{Defining Attack Success}
The goal of object detectors is to identify and locate objects in the input image with 2D boxes, which we refer to as $\{b\}^{(o)}_i$ when they belong to a legitimate image and $\{b\}^{(c)}_i$ when they belong to the attack-corrupted image.
Figure~\ref{fig:object_detection_vis} shows an example of the attack effect on the detected objects: in the rolling shutter attack-corrupted image, many objects are mis-detected compared to a clean input.
In practice, to compute the attack effect, for each box, we categorized the effect of the attack on each object box based on the Intersection-over-Union (IoU) between original and corrupted boxes.
IoU is a measure of the overlapping area of two boxes, as a proportion of the total, combined area; it is commonly used to evaluate the accuracy of a box prediction compared to a ground truth box.
We measured the attack outcome with the effect that the distortions have on the detected boxes with the following:
\begin{itemize}[leftmargin=.5cm]
	\item \textbf{Hidden}: for a box $b^{(o)}_i$ its object has been \textit{hidden} if there is no box in the corrupted image with IoU($b^{(o)}_i, b^{(c)}_j$) > 0.5. 
	\item \textbf{Misplaced}: for a box $b^{(o)}_i$ its object has been \textit{misplaced} if there is a box in the corrupted image with IoU smaller than 0.95 and same output class. 
	\item \textbf{Appeared}: for a box (in the corrupted image) $b^{(c)}_j$, its object has \textit{appeared} if there is no box in the original image with IoU($b^{(c)}_j, b^{(o)}_i$) > 0.5.
\end{itemize}

\subsubsection{Method}

\paragraph{Attack Pattern Collection}\label{sec:coco_sim_setup}
We simulated the attack to find the best configuration of adversary-controlled parameters (i.e., $t_{on}$ and $f$): the one that leads to the highest number of hidden objects.
We collected the patterns as described in Section~\ref{sec:exp-setup}.
Given that the Axis camera's frame rate is 25 fps ($F$=25), we collected patterns for four different laser frequencies, $f = 25\text{, }250\text{, }500\text{, and } 750\text{Hz}$.
Since the image sensor of the Axis camera has dead areas, the number of distortions can be calculated using Equation~\ref{eq:number_of_bars_dead_area}.
We collected patterns with duty cycles from 0.1\% to 40\% (from 1.3$\mu s$ to 533.3$\mu s$), because a change in $f$ or duty cycle leads to different $t_{on}$ (see Equation~\ref{eq:on_time}).
It should be noted that $t_{on}$ refers to the time the laser is on per-distortion rather than per-frame.
Since $t_{exp}$ affects the attack but is not controlled by the attacker --- as it is set by the auto-exposure mechanism --- we tested empirically which $t_{exp}$ the camera would set on a sunny or cloudy day outdoors, finding 32$\mu$s and 200$\mu$s, respectively.
We used this range of $t_{exp}$ to collect the patterns.

\paragraph{Simulation Setup}

We simulated the attack effect on a subset of two different video datasets, namely BDD100K~\cite{yu2020bdd100k}, a large and diverse driving video dataset, and VIRAT~\cite{oh2011large}, a dataset of surveillance footage.
We randomly picked 50 videos from the BDD100K and, due to the longer duration, 25 videos from the VIRAT dataset, for each video, we extracted every 10$^{\text{th}}$ frame.
For each parameter configuration ($f, t_{exp}, t_{on}$), we randomly selected ten patterns.
Next, we generated approximately 7 million corrupted frames by overlaying the extracted patterns over the legitimate video frames.
Finally, we assessed the simulated attack outcome by comparing legitimate and corrupted frames, as showcased in Figure~\ref{fig:object_detection_vis}.
We evaluated two well-known state-of-the-art object detection models from the Tensorflow Model Zoo, namely Single Shot Detector (SSD)~\cite{liu2016ssd} and Faster RCNN~\cite{ren2015faster} (FRCNN), both using Inception v2 backbone network.
For performance reasons, we resized all frames to 640$\times$360 pixels, before they were passed into the object detectors.\footnote{Implementation and evaluation code is available at \url{https://github.com/ssloxford/they-see-me-rollin}}

\subsubsection{Results}
\begin{figure}[t]
	\centering
	\includegraphics[width=0.95\linewidth]{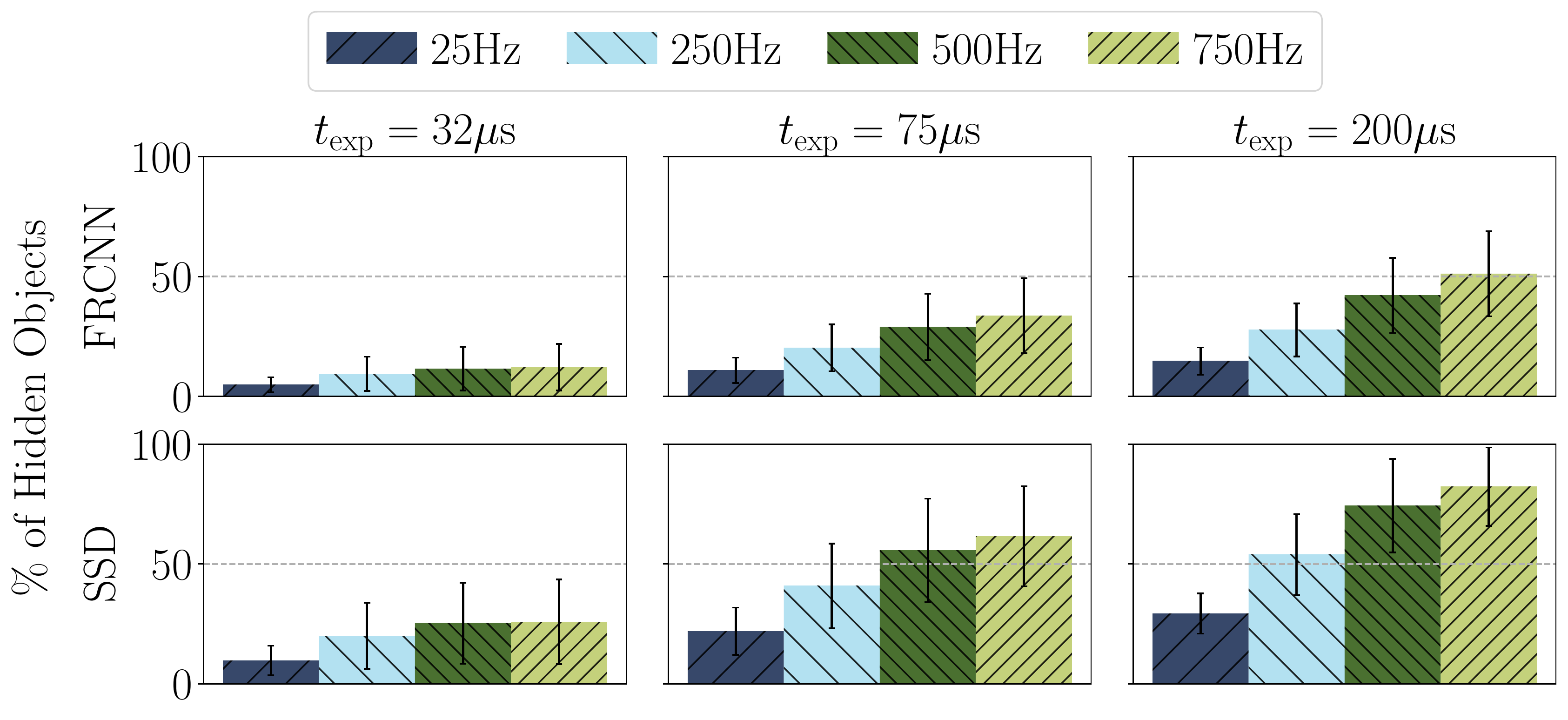}
	\caption{Percentage of hidden objects in the two video datasets for $f = 25\text{, }250\text{, }500\text{, and } 750\text{Hz}$, various $t_{exp}$ and the two object detection models. Error bars show the standard deviation over the results per-video.}
	\label{fig:res_choose_freq} 
\end{figure}

\paragraph{Choosing Attack Parameters}

The percentage of objects hidden in the input frames for the various frequencies and exposure times is presented in Figure~\ref{fig:res_choose_freq}.
It should be noted that for all four frequency settings, the image area covered by the laser-injected light is the same.
As the frequency increases, $t_{on}$ decreases, which results in smaller, but more frequent distortions (see Equation \ref{eq:on_time}).
However, the sum of $t_{on}$ per frame stays the same.
Figure~\ref{fig:res_choose_freq} shows a clear increment in the ratio of hidden objects when using larger $f$. 
For $f=750$Hz, we obtained the largest number of hidden objects.
At $t_{exp}=200\mu$s, 51\% and 82\% of objects were hidden for FRCNN and SSD, respectively.
We noted that the narrower distortions generated by higher frequencies strongly affect the object detection task, in particular, for small objects.
Given these results, an adversary would select 750Hz as the modulation frequency for the attack, as this selection better generalizes to different input scenes.
In line with this, we limit the presentation of the remaining results to this 750Hz configuration.

\paragraph{Effect of Exposure}
\begin{figure}[t]
	\centering
	\includegraphics[width=0.89\linewidth]{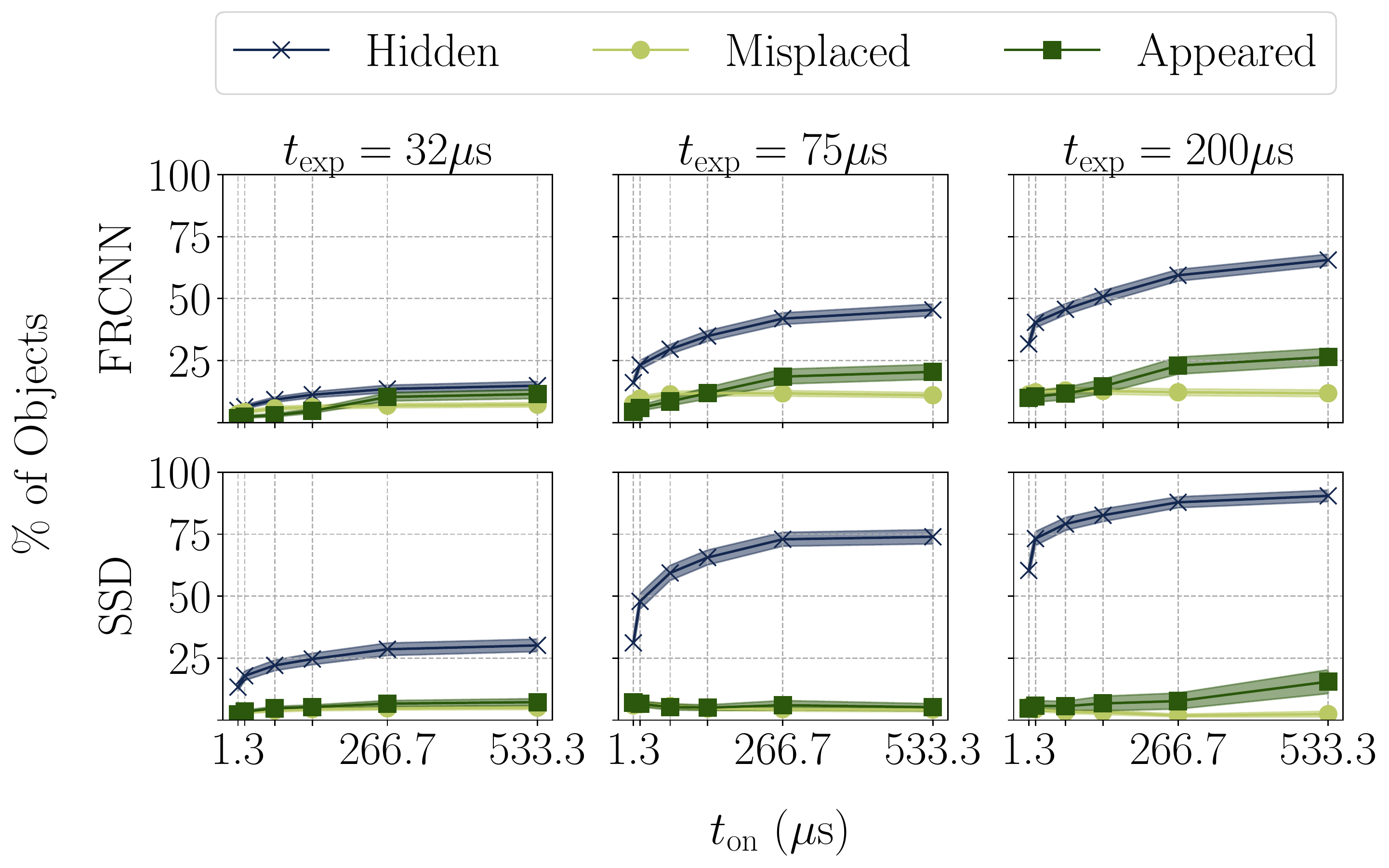}
	\caption{Percentage of hidden, misplaced, and appeared objects in the evaluation of the BDD100K and the VIRAT dataset, for various $t_{on}$, exposure times $t_{exp}$, and the two models used in the evaluation (SSD and FRCNN). Shaded areas show 99\% confidence intervals.} 
	\label{fig:res_effect_exposure} 
\end{figure}

As the exposure time is not controlled by the adversary we investigated the effect of $t_{exp}$.
Figure~\ref{fig:res_effect_exposure} shows the ratio of hidden, misplaced, or appeared objects for various exposure times $t_{exp}$ and laser on-times $t_{on}$.
With increasing $t_{exp}$ and thus greater distortion, more objects are misplaced or hidden.
As expected, an increasing $t_{on}$ also leads to more objects being hidden. 
However, the number of misplaced or appeared objects does not change significantly.
In contrast, for short $t_{exp}$ less of the injected light is absorbed, leading to less intensive distortions and fewer hidden objects.
While the weak illumination could be compensated with a more powerful laser, this highlights how in dimmer ambient light settings, the attack requires fewer resources on the adversary's side.
For the largest evaluated exposure ($t_{exp}=200\mu$s) and $t_{on}=533\mu$s, our attack can hide up to 90\% of objects in an image for SSD and up to 65\% for FRCNN.
This is in line with the fact that SSD is designed with more focus on speed of execution rather than prediction accuracy: SSD can execute one inference every 30ms, while FRCNN needs 89ms.

The impact that the rolling shutter attack can have on an autonomous driving platform as a whole, where object detection is used together with other components, is illustrated in Appendix~\ref{sec:pylot_evaluation}.

\begin{figure}[t]
	\centering
	\includegraphics[width=0.99\linewidth]{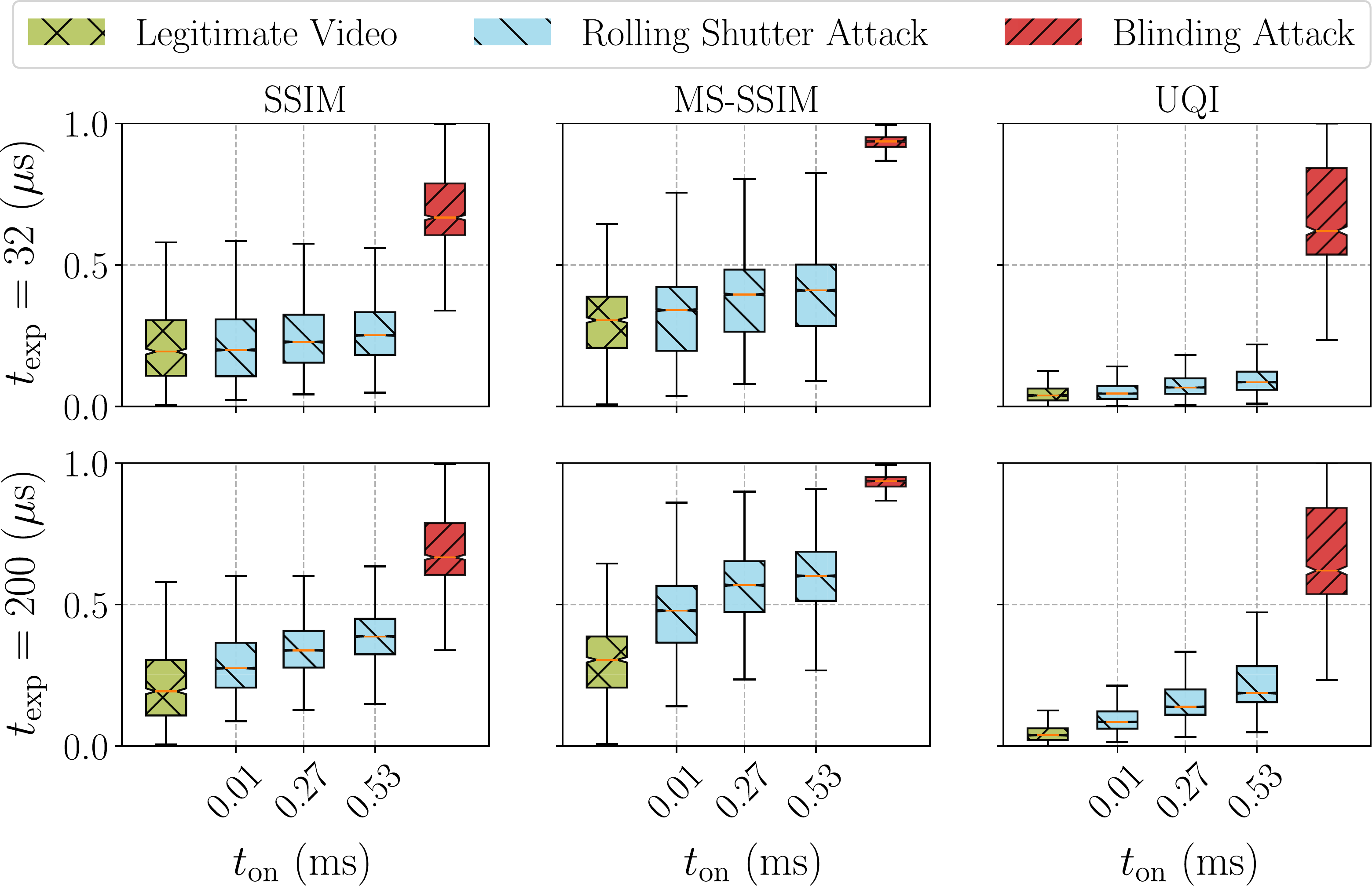}
	\caption{Image perturbation metrics between consecutive frames for $f$ = 750Hz during normal operation, a rolling shutter attack and a blinding attack. High values correspond to large perturbations. 
	} 
	\label{fig:res_image_metrics} 
\end{figure}

\subsection{Comparison with Blinding Attack}\label{sec:comparison_with_blinding}

\subsubsection{Method}

To measure the perturbation caused by our attack, we repeated parts of the simulation on the BDD100K dataset as described in Section~\ref{sec:evaluation_setup}. 
We randomly selected 100 videos from the \texttt{validation} set and extracted 10 evenly spaced frame pairs, whereas each pair consisted of two consecutive frames.
For clarity, we refer to those frames as \textit{previous} and \textit{current}.
We then calculated the amount of perturbation between those consecutive frames to evaluate how the \textit{current} frame has changed compared to the \textit{previous} frame during (i) normal operation, (ii) a rolling shutter attack, and (iii) a blinding attack.
For (ii) and (iii), we generated corrupted frames by overlaying the malicious patterns, as described in Section~\ref{sec:evaluation_setup}, onto the \textit{current} frame of each pair.
We measured the image perturbation with several metrics from adversarial machine learning literature~\cite{sharif2018suitability}.
For each scenario and pair of frames, we calculated three different metrics: the Structural Similarity (SSIM), Multi-Scale SSIM (MS-SSIM) and Universal Image Quality Index (UQI).
The values for SSIM, MS-SSIM and UQI are bounded in [0, 1] and were inverted in our case to represent \textit{dissimilarity} rather than similarity. 
In addition, we removed perturbations with a value of 0, as these correspond to stationary frames.
To measure the detection performance for pairs of anomalous frames, we set up a simple binary-threshold approach, i.e., if the dissimilarity metric value is higher than a certain threshold, we label the frame pair as anomalous.
We report detection results with the area under the receiver operating characteristic curve (ROC-AUC), where a score of 1.0 corresponds to perfect detection while a score of 0.5 corresponds to random guess.

\subsubsection{Results}

Figure~\ref{fig:res_image_metrics} shows the frame-to-frame perturbation introduced by the blinding and the rolling shutter attack for $t_{exp}$ = 32$\mu$s and $t_{exp}$ = 200$\mu$s.
The results indicate that for short $t_{exp}$ our attack is causing interference that is similar across all metrics to the expected level of perturbation seen in consecutive legitimate video frames.
In contrast, the blinding attack generates much larger perturbations compared to our attack, which are easily detected.
We found that SSIM, MS-SSIM and UQI could detect almost 100\% of blinding attacks with the binary-threshold approach, i.e., ROC-AUC of 0.99, 1.0 and 1.0, respectively.
While higher $t_{exp}$ causes more perturbation, the level of interference is still closer to that for legitimate video frames than for blinding.
In comparison, for most cases our attack was indistinguishable from legitimate frame-to-frame changes when using the same threshold approach.
The ROC-AUC scores for the detection of the rolling shutter attack with $t_{exp}$ = 32$\mu$s and $t_{on}=0.53$ms are 0.61, 0.68 and 0.80, for SSIM, MS-SSIM and UQI, respectively.
Similar results (0.67, 0.79, 0.81) were observed for $t_{exp}$ = 200$\mu$s and $t_{on}=0.01$ms.
However, as expected, the attack is more easily detected with increasing $t_{on}$.
Nevertheless, choosing the worst performing scenario for the rolling shutter attack at the 0\% false positive rate threshold ($t_{on}=0.53$ms, $t_{exp}=32\mu$s and using MS-SSIM), we obtain that 100\% of blinding attacks are detected while only 6.55\% of rolling shutter attacks are.

\section{Countermeasures}

The vulnerability that enables the rolling shutter attack is inherent to CMOS image sensors, meaning that attacks cannot be prevented entirely without changing the hardware design. 
However, detection in software can mitigate the attacks' impact, while other countermeasures are applicable in some special cases. 

\subsection{Hardware Countermeasures}

A clear design option is to use a camera with a global shutter mechanism.
This does not stop the injected light but prevents rolling shutter attacks as the image rows are exposed at the same time, leading to a simple blinding attack. 
However, this requires a design-time choice and also carries additional component costs for incorporating frame memory or using CCD image sensors.
These costs increase dramatically for existing systems, as the replacements must be retrofitted as well. 

Additionally, camera redundancy is a straightforward protection approach, but also quickly increases manufacturing costs for little improvement in security. 
In some scenarios, the existing use of multiple sensors allows the system to continue operating effectively despite the performance degradation caused by the rolling shutter attack (e.g., in autonomous driving, with multiple cameras or LIDAR). 
However, it is not trivial to reliably determine which sensor input is malicious or untrustworthy in order to stop relying on it. 
Furthermore, while a system may be tolerant to the degradation of its sensor inputs, there is still some finite budget for degradation that it can withstand before it becomes unsafe or ineffective. 

\subsection{Deep Learning Based Detection}

Here, we explore the re-use of the deep network used for the computer vision task to detect the presence of the attack. 
We re-use the backbone network using the insight that appearance of the distortions can be captured by early convolutional filters, designed to capture low-level input features.

\paragraph{Setup and Training}

Given the backbone network, we added a network head as follows:
on top of the first convolutional layer (after pooling, when present) we added one additional convolution layer with 32 filters, then we used global maximum pooling and added a binary classification layer which predicts whether the input is corrupted by a rolling shutter attack.
We also used dropout and selected leaky ReLu as the activation function, which has been found to lead to faster convergence and improved performance~\cite{xu2015empirical}.
We created a dataset to train our detection by splitting both collected patterns and video frames into training, validation and testing subsets, with ratios 60\%, 20\% and 20\%, respectively.
As in Section~\ref{sec:evaluation_setup}, corrupted frames were obtained by overlaying the pattern onto the legitimate frame.
We used all patterns collected for the Axis camera ($\sim$6,500, see Section~\ref{sec:exp-setup}) and 180 videos randomly selected from BDD100K, which we sampled every 10$^\text{th}$ frame (20,880 frames).
We used cross entropy loss and trained the added parameters with Adam, with learning rate $10^{-4}$ and for 10 epochs.

\paragraph{Results}

\begin{table}[t]
	\centering
	\caption{Rolling shutter attack detection performance for the backbone networks. 
	False Negatives (FN) and False Positives (FP) are calculated over 3,560 test frames.
	The column \textit{Overhead} represents the computational overhead for the detection mechanism in percentages with the standard deviation in parentheses.}
	\small
	\begin{tabular}{ccccccc}
		\toprule
		\textbf{Model} & \textbf{Backbone} & \textbf{Accur.} & \textbf{FN} & \textbf{FP} & \textbf{Overhead in \%} ($\pm \sigma$)  \\
		\midrule
		\textit{SSD} & \textit{Mobilenet v2} & 0.98820 & 32 & 10 & +0.57 ($\pm$4.059) \\
        \textit{SSD} & \textit{Inception v2} & 0.99494 & 18 & 0 & +0.13 ($\pm$4.242) \\
        \textit{FRCNN} & \textit{Inception v2} & 0.99551 & 7 & 9 & -0.11 ($\pm$5.045) \\
        \textit{FRCNN} & \textit{Resnet-50} & 0.98989 & 25 & 11 & +1.38 ($\pm$5.119) \\
\bottomrule
	\end{tabular}
	\label{tab:detection_accuracy}
\end{table}

The test set accuracies for the two object detectors with different backbones are presented in Table~\ref{tab:detection_accuracy}.
The detection is efficient, all networks obtain over 98.8\% accuracy on the task, although MobileNet v2 performs slightly worse due to design compromises for faster execution speed.
Table~\ref{tab:detection_accuracy} also reports the number of False Negatives (FN) and False Positives (FP). 
This shows that setting the detection threshold to 0.5 would generally tend to favor low false positives, but a different trade-off can be sought by changing the threshold for detection.

To analyze the performance impact of the proposed detection mechanism, we measured the required inference time for all images in the test set (batch size = 1) with and without the added layers on an NVIDIA TITAN RTX.
The last column of Table~\ref{tab:detection_accuracy} shows the computational overhead of the detection, indicating that it is negligible.
As such, this detection measure could be applied as a software patch to protect existing systems. 
Additionally, the method represents a general approach that can be applied to protect any computer vision network from rolling shutter attacks. 

\section{Limitations}

In this section, we provide an overview of the limitations of our evaluation and the rolling shutter attack itself.

\subsection{Evaluation Limitations}

The evaluation presented in this paper is a first step in raising awareness about the rolling shutter attack and its impact.
Nevertheless, our attack evaluation has some limitations that an attacker might need to consider when performing the attack.
Although the patterns we used have been collected from real cameras under close to real-world settings, our simulations do not perfectly reflect the dynamic behavior of the target camera under attack.
Depending on the camera, numerous factors and mechanisms may influence the appearance of the injected distortion.
We mimicked the behavior by exhaustively simulating the attack with different patterns that correspond to different exposure times.
Nevertheless, depending on the scenario, an attacker might not always freely choose parameters $f$ and $t_{on}$ while maintaining the desired brightness of the injected distortions.
For instance, in bright settings, short $t_{on}$ and non-powerful lasers might not generate enough light to be captured by the camera.
Furthermore, we have not explicitly considered the effect of auto-exposure and auto-focus mechanisms.
While we found that these are generally not triggered for most of our attack configurations, we cannot exclude that these mechanisms might come into play as the attack is executed (this will depend on camera-specific factors).
This will have an effect on the appearance of the distortions and may cause warnings in the vision-based system.

\subsection{Attack Limitations}

Due to the characteristics of the rolling shutter attack, there are some limitations that cannot be easily solved.

\paragraph{Limitations in Dim Conditions}
Contrary to what humans perceive, the amount of ambient light does not scale linearly between what is perceived as bright versus what is perceived as dark.
For example, on a clear day, the amount of ambient light at midday is >30k lx, at sunset, it is $\sim$400 lx while at night it is close to 0 lx.
Cameras are no exception: in poor lighting conditions, cameras have to increase the exposure time \textit{dramatically} to capture enough light.
This leads to a large overlap of consecutive pixel rows resulting in fine-grained distortions not being achievable.
Therefore, in dim conditions, even a very short $t_{on}$ will impact a significant proportion of image rows, limiting the amount of high-frequency distortions from which the attack benefits.

\paragraph{Timing \& Synchronization}
While it is, in principle, possible to target specific rows during the attack, tight synchronization between the modulation signal and the row reset signal of the image sensor is required.
Altering the content of the $i$-th row to hide a specific object, requires exact knowledge about the exposure timing.
In case the attacker has access to the video feed, such synchronization could be obtained through an iterative process.
However, as in our threat model, it is generally reasonable to assume that the attacker does not have this information and therefore needs to design their attack accordingly.

\paragraph{Frame Rate Accuracy}
To control the movement of distortions between consecutive frames, the laser modulation must vary in relation to the camera frame rate, or a multiple of it. 
Accurate knowledge of the frame rate allows the attacker to keep the progression of the pattern in line with their intentions. 
The frame rate $F$ of a camera is typically specified in the datasheet, although minor timing inaccuracies might cause the frame rate to deviate slightly (e.g., 30.0025 fps instead of 30 fps). 
However, this only causes the pattern to drift slowly if the attack is carried out over an extended period of time and is negligible for attacks of short duration.

\section{Conclusion}

In this paper, we have highlighted an inherent vulnerability of the electronic rolling shutter mechanism implemented in most modern CMOS image sensors, which can be exploited to inject fine-grained distortions into the captured frames. 
We showed how an adversary can construct an attack in practice by introducing a model to accurately predict the size of the injected distortions. 
We describe how this model can be used to account for uncontrolled camera-specific and environmental variations, in order to make the attack more reliable. 
Using only off-the-shelf equipment, we reproduced the attack steps in practice on two separate cameras.
We then investigated the attack within the context of an object detection task, where the camera-captured images are used by computer vision models to locate and classify objects.
Based on two well-known video datasets, we evaluated the performance of two state-of-the-art object detection models under the rolling shutter attack, showing that for SSD, our attack can hide up to 75\% of objects present in the input frames.
At the same time, we showed that our attack is stealthier than a global blinding attack across several anomaly metrics, making it harder to detect.
Our findings outline a weakness in using cameras with CMOS images sensors in settings with potentially-adversarial camera inputs, and particularly for computer vision applications in safety-critical systems. 

\begin{acks}

This work was supported by a grant from Mastercard and by the Engineering and Physical Sciences Research Council (EPSRC). 

\end{acks}

\bibliographystyle{ACM-Reference-Format}
\bibliography{references}

\balance

\clearpage

\appendix
\renewcommand{\thesection}{\Alph{section}.\arabic{section}}
\setcounter{section}{0}
\nobalance

\begin{appendices}

\section{Estimating $t_{exp}$}\label{sec:estimate_exp}

On the basis of the Axis M3045-V, we demonstrate how the adversary can estimate the exposure time $t_{exp}$.
Moreover, we show that the method leads to very accurate results, by comparing frames captured with auto-exposure and manually set exposure.

\paragraph{Method}

In the first step, we extracted $H_{v} = 0.25$lx from the datasheet.
We then collected multiple frames with the auto-exposure mechanism.
Subsequently, we deactivated the auto-exposure mechanism, measured the current ambient light level $E_v$ with a lux meter and used Equation~\ref{eq:estimate_texp} to estimate $t_{exp}$.
We repeated the frame collection for different lighting conditions, such as at sunset, on a cloudy day and a sunny day.

\paragraph{Accuracy of Estimation}

\begin{table}[]
    \centering
    \small
	\caption{Results of the cross-correlation between the color channel histograms of the frames captured with auto-exposure, manual exposure setting and manual exposure setting with an offset of 60$\mu$s.}
    \begin{tabular}{ccc}
    	\toprule
        \textbf{Color Channel} & \textbf{Manual} & \textbf{Manual + Offset} \\\hline
        Red             & 0.9961 & 0.9858          \\
        Green             & 0.9986 & 0.9865          \\
        Blue             & 0.9939 & 0.9545          \\\bottomrule
    \end{tabular}
    \label{tab:cross_corr_histograms}
\end{table}

On a first glance, there was no optical difference between the frames collected with auto-exposure and the ones collected with manual settings. 
To analyze the accuracy, we calculated histograms for each color channel (R, G and B) of the collected frames.
We then calculated the cross-correlation between the color histograms of the frames captured with auto-exposure and the ones with manually set exposure.
The results in Table~\ref{tab:cross_corr_histograms} show that there is only a small difference between the frames captured with auto-exposure and the ones with manual exposure.
To make the results more comparable, we also collected frames with an exposure of $t_{exp} + 60\mu$s.
We picked an offset of 60$\mu$s, because this was the first value where we noticed considerable changes in brightness.
As can be seen in Table~\ref{tab:cross_corr_histograms}, an offset of 60 also led to a noticeable difference in the cross-correlation of all three color channels.
This suggests that the estimation of $t_{exp}$ is accurate.

\section{Incorrect $t_{exp}$ - Additional Results}\label{sec:estimate_texp_incorrect}

\begin{figure}[t]
  \centering
  \includegraphics[width=.90\linewidth]{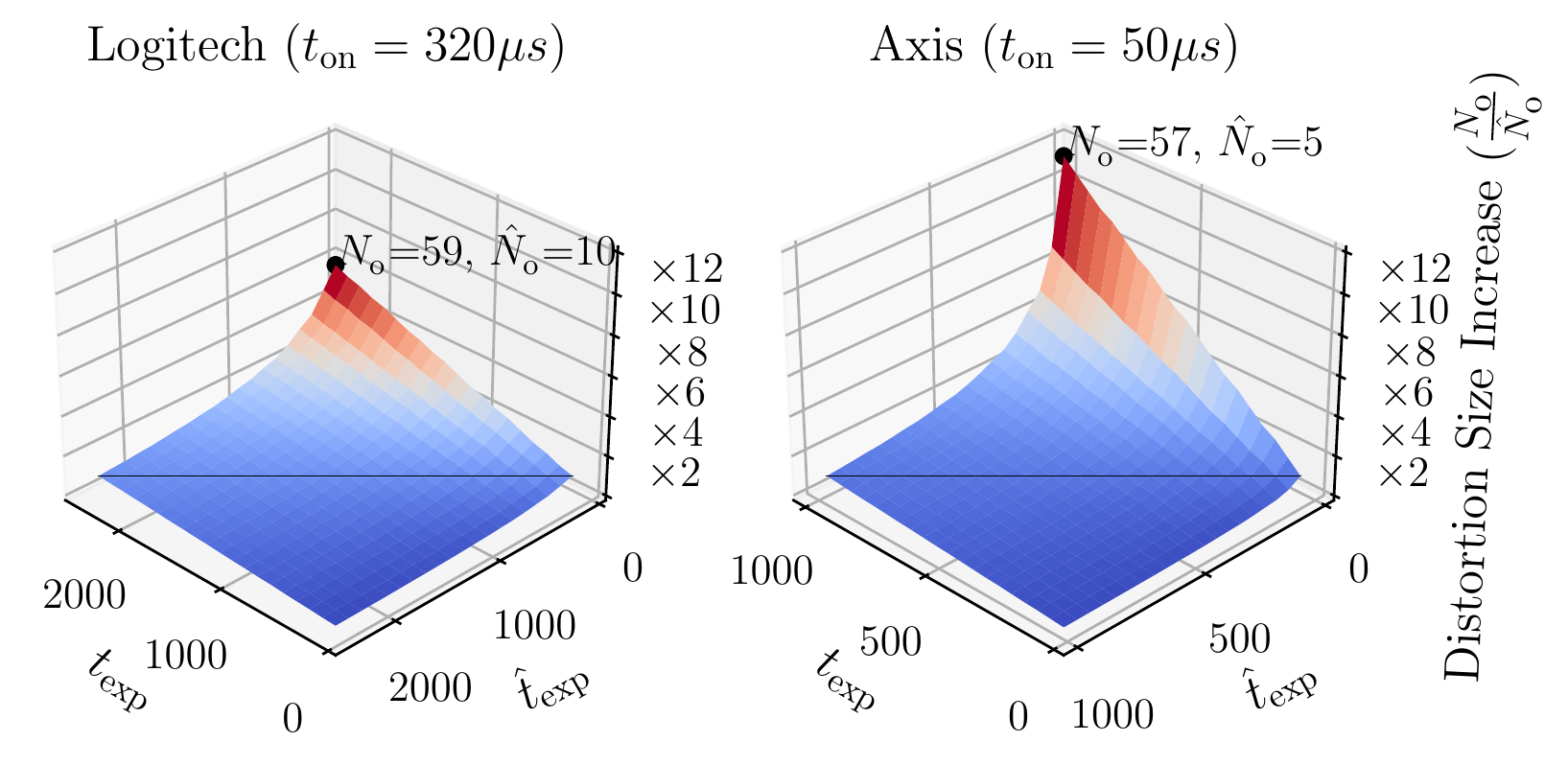}
  \caption{As in Figure~\ref{fig:sec6-incorrect-texp}, increment in the expected distortion size $\hat{N}_{o}$ and actual size $N_{o}$ as the adversary's exposure time estimate $t_{exp}$ diverges from the true exposure time value $\hat{t}_{exp}$. Shortest $t_{on}$ used in the data collection of Section~\ref{sec:rsmeval}.} 
  \label{fig:enough1} 
\end{figure}

\begin{figure}[t]
  \centering
  \includegraphics[width=.90\linewidth]{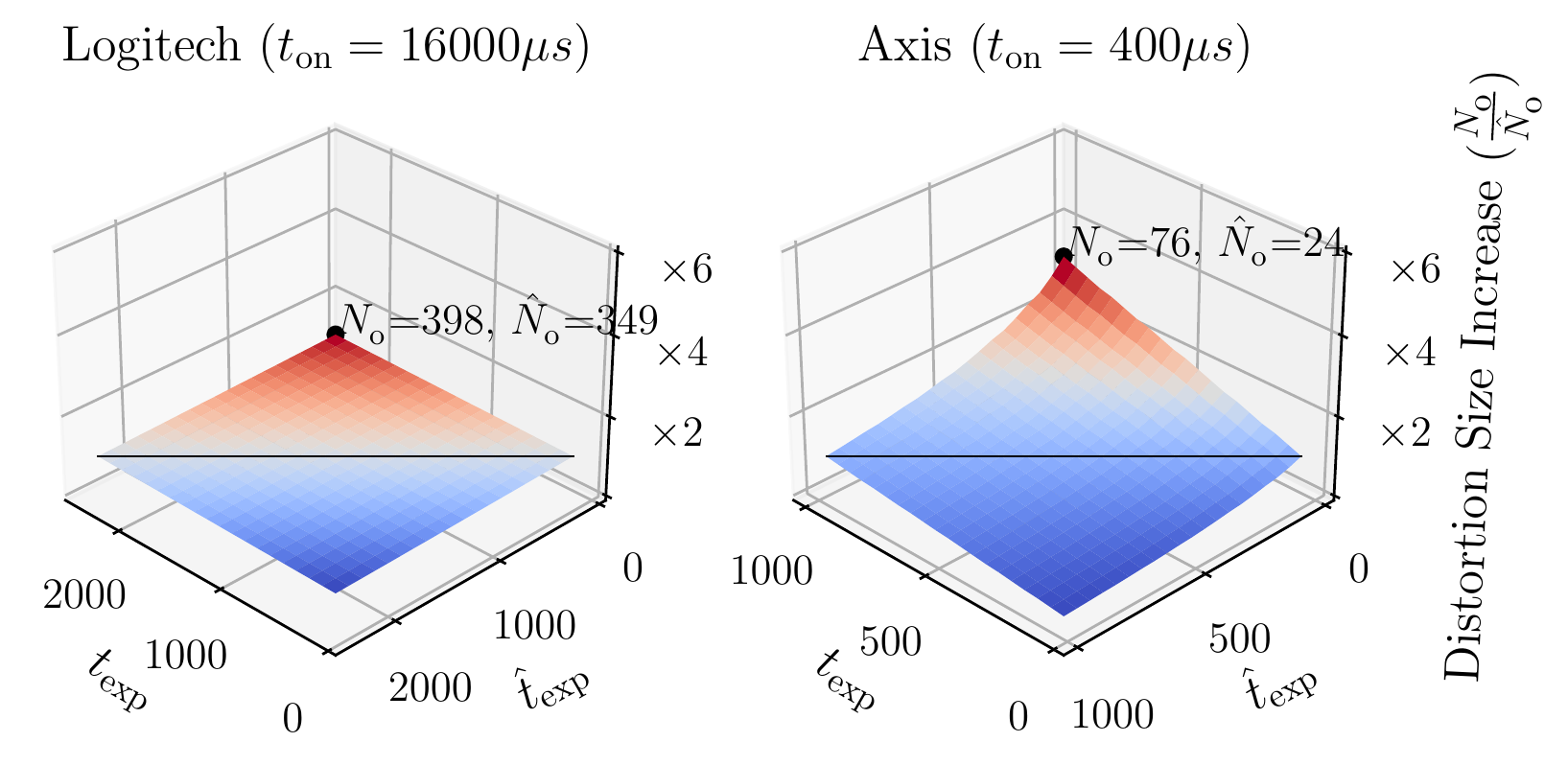}
  \caption{As in Figure~\ref{fig:sec6-incorrect-texp}, increment in the expected distortion size $\hat{N}_{o}$ and actual size $N_{o}$ as the adversary's exposure time estimate $t_{exp}$ diverges from the true exposure time value $\hat{t}_{exp}$. Longest $t_{on}$ used in the data collection of Section~\ref{sec:rsmeval}.} 
  \label{fig:enough2} 
\end{figure}

As reported in Section~\ref{sec:acc_of_shutter_m}, incorrect estimates of the exposure time can lead to distortion sizes that diverge from the expected size. 
Here we report the best and worst divergence observed for the parameters used in our data collection, which depends on $t_{on}$.
In fact, larger $t_{on}$ will lead to smaller variation, and vice-versa.
The shortest $t_{on}$ is reported in Figure~\ref{fig:enough1}, the longest $t_{on}$ is reported in Figure~\ref{fig:enough2}. 
The figures show that for larger $t_{on}$ (which also generally produce better attack success in our evaluation, Section~\ref{sec:attacking-object-detection}), incorrect estimates tend to lead to diminishing divergence in expected versus actual distortion size.
For Logitech we get a worst-case distortion size increase of 1.6 (down from 6 in the shortest $t_{on}$ setting), while the same increase for Axis goes from a factor of 11 (in the shortest $t_{on}$ setting) down to a factor of 3.

\section{Effect on Autonomous Driving}\label{sec:pylot_evaluation}

We used the open-source simulator CARLA~\cite{Dosovitskiy17}, in combination with the self-driving car platform Pylot~\cite{gog2021pylot} to illustrate the effects of the rolling shutter attack on an autonomous system.
By default, the Pylot agent is equipped with four sensors: (i) a wide-angle RGB camera for lane detection and object detection/tracking, (ii) a telephoto camera for traffic light detection, (iii) a LiDAR sensor for localization, and (iv) GPS for route planning.
We uniquely targeted the wide-angle RGB camera with our attack, to measure how the system as a whole is affected.

\paragraph{CARLA Setup}
We utilized two predefined scenarios from the CARLA scenario runner, where an entity is crossing the road in front of the car: \textit{Scenario 1}  involves a pedestrian and \textit{Scenario 2} involves a cyclist.
We collected the baseline performance of Pylot by simulating each scenario 50 times.
We then repeated the simulation under the presence of the rolling shutter attack.
We intercepted the frames captured by the wide-angle camera as they were forwarded to Pylot and overlayed the rolling shutter pattern, as described in Section~\ref{sec:coco_sim_setup}.
The attack parameters were as follows: $f=750$Hz, $t_{exp}=200\mu$s and $t_{on}=0.53$ms.
In line with the previous evaluation, we again used two object detectors, FRCNN and SSD, which are among the pre-implemented options in Pylot.

\paragraph{Results}

\begin{figure}[t]
	\centering
	\includegraphics[width=1\linewidth]{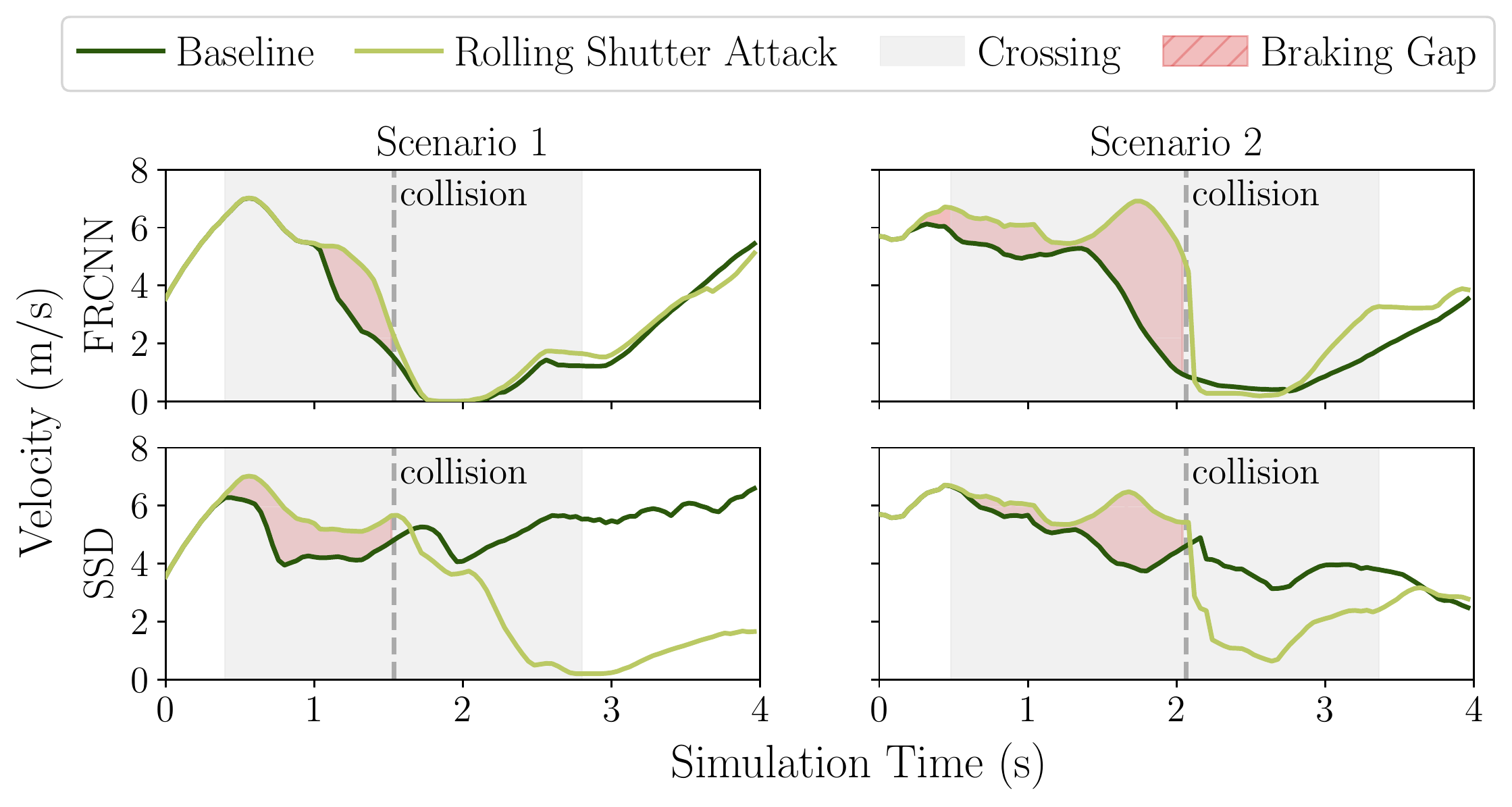}
	\caption{Velocity curve of the agent during normal operation (baseline) and during the rolling shutter attack. The grey area shows the time when the pedestrian/cyclist crosses the road. The dashed line marks the time when the agent collides with the non-player character.} 
	\label{fig:pylot_velocity_graphs}
	\vspace{-0.1cm} 
\end{figure}

Most of the key findings from Section~\ref{sec:attacking-object-detection} can be transferred to the CARLA simulations.
In particular, the performance difference between SSD and FRCNN stands out.
While for \textit{Scenario 1} during normal operation for both object detectors the probability of a collision was equal (5\%), under the rolling shutter attack, SSD showed a 67\% probability of a collision compared to 47\% for FRCNN.
Similar behavior can be observed for the baseline of \textit{Scenario 2}.
FRCNN showed a 0\% probability of a collision, while using SSD led to an increase to 15\%.
However, interestingly, under attack, FRCNN experienced a strong performance drop and fell behind SSD.
The probability of a safety infraction increased to 97\% compared to 65\% for SSD.

In Figure~\ref{fig:pylot_velocity_graphs}, the velocity curve of the autonomous vehicle for the two scenarios and object detectors is presented.
The figure shows that in the baseline, the car immediately reduces speed as soon as the non-player character starts moving.
In contrast, as the braking gap shows, during the attack, a noticeable delay in braking can be observed.

\begin{figure*}[t]
	\centering
	\begin{subfigure}[b]{0.49\textwidth}
		\centering
		\includegraphics[width=.98\textwidth]{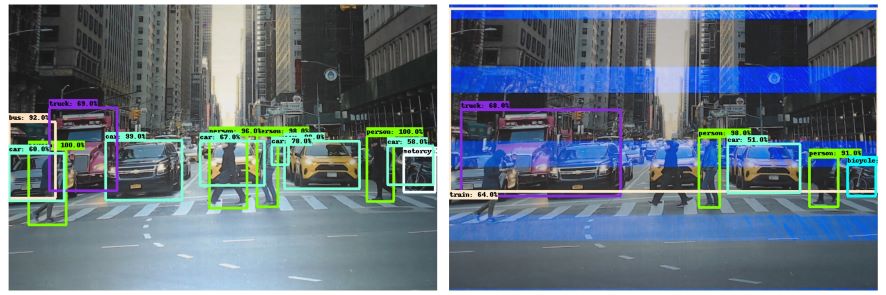}
		\label{fig:frcnn_cam_logitech}
		\caption{Logitech C922}
	\end{subfigure}
	\begin{subfigure}[b]{0.49\textwidth}
		\centering
		\includegraphics[width=.98\textwidth]{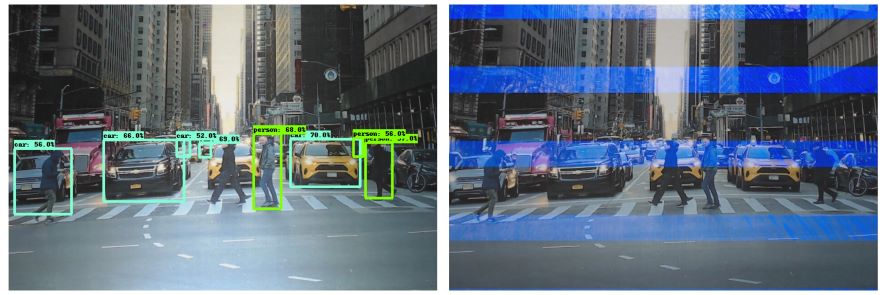}
		\label{fig:ssd_cam_logitech}
		\caption{Logitech C922}
	\end{subfigure}
	\begin{subfigure}[b]{0.49\textwidth}
		\centering
		\includegraphics[width=.98\textwidth]{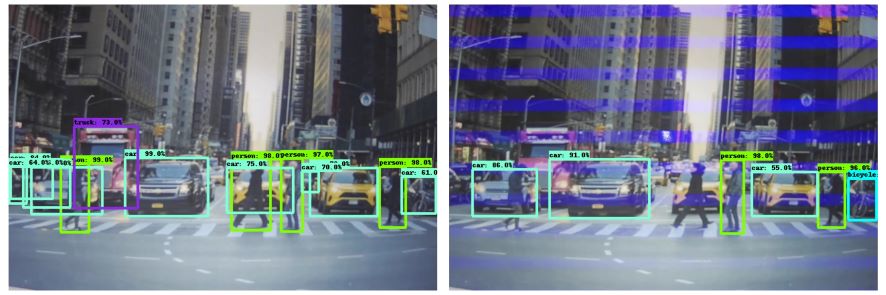}
		\label{fig:frcnn_cam_axis}
		\caption{Axis M3045-V}
	\end{subfigure}
	\begin{subfigure}[b]{0.49\textwidth}
		\centering
		\includegraphics[width=.98\textwidth]{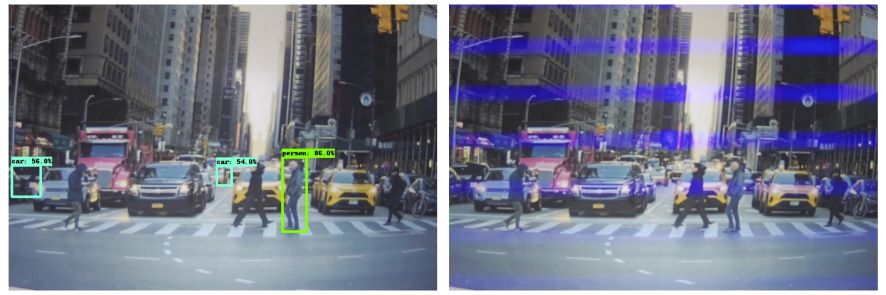}
		\label{fig:ssd_cam_axis}
		\caption{Axis M3045-V}
	\end{subfigure}
	\begin{subfigure}[b]{0.49\textwidth}
		\centering
		\includegraphics[width=.98\textwidth]{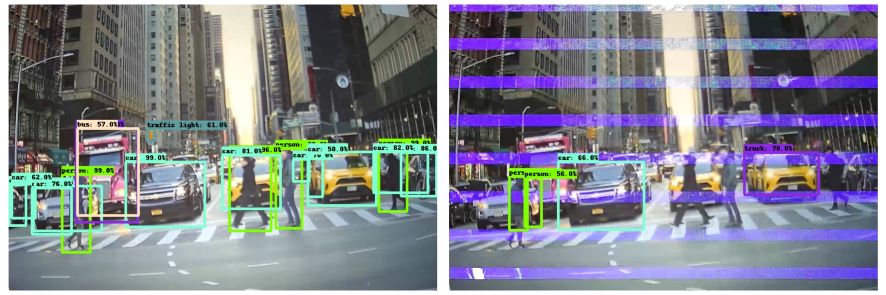}
		\label{fig:frcnn_cam_yi}
		\caption{YI Home Camera 1080p}
	\end{subfigure}
	\begin{subfigure}[b]{0.49\textwidth}
		\centering
		\includegraphics[width=.98\textwidth]{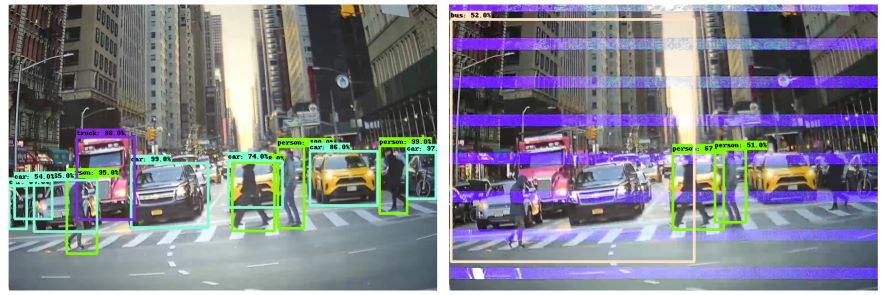}
		\label{fig:ssd_cam_yi}
		\caption{YI Home Camera 1080p}
	\end{subfigure}
	\begin{subfigure}[b]{0.49\textwidth}
		\centering
		\includegraphics[width=.98\textwidth]{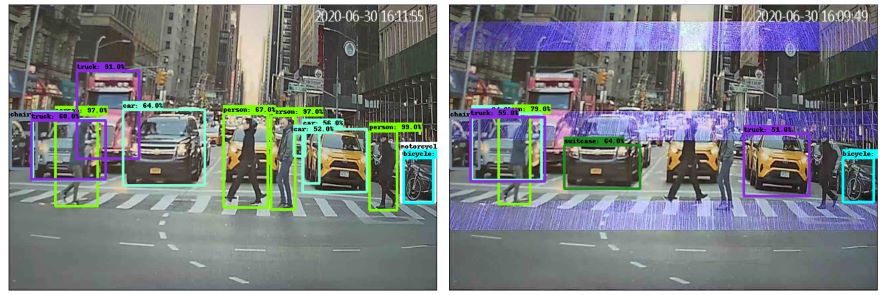}
		\label{fig:frcnn_cam_v380}
		\caption{V380 Camera 720p}
	\end{subfigure}
	\begin{subfigure}[b]{0.49\textwidth}
		\centering
		\includegraphics[width=.98\textwidth]{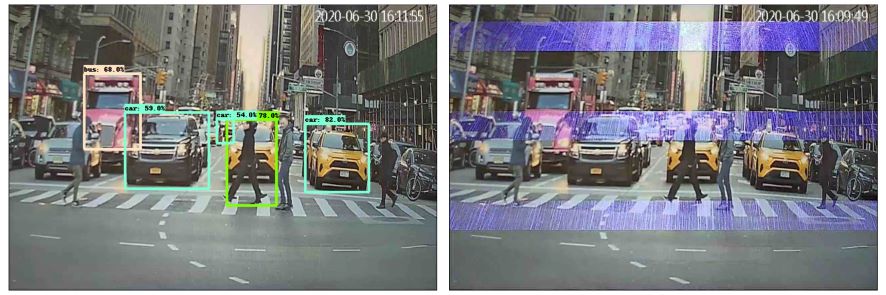}
		\label{fig:ssd_cam_v380}
		\caption{V380 Camera 720p}
	\end{subfigure}
		\begin{subfigure}[b]{0.49\textwidth}
		\centering
		\includegraphics[width=.98\textwidth]{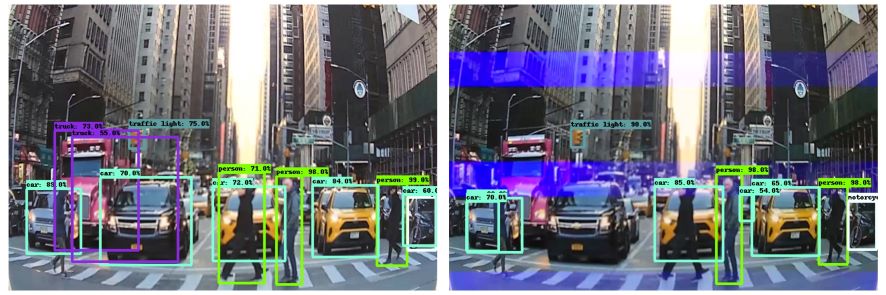}
		\label{fig:frcnn_cam_arlo}
		\caption{Netgear Arlo}
	\end{subfigure}
	\begin{subfigure}[b]{0.49\textwidth}
		\centering
		\includegraphics[width=.98\textwidth]{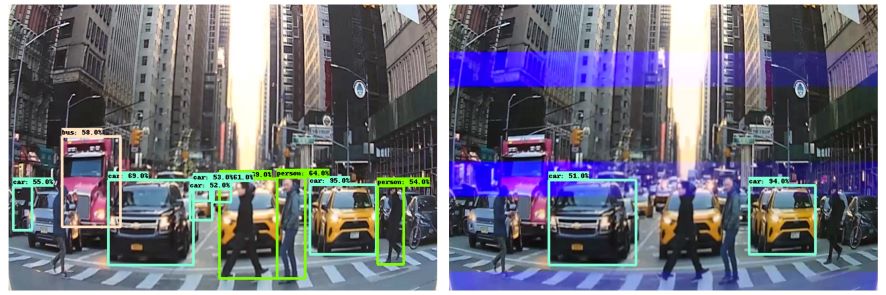}
		\label{fig:ssd_cam_arlo}
		\caption{Netgear Arlo}
	\end{subfigure}
	\begin{subfigure}[b]{0.49\textwidth}
		\centering
		\includegraphics[width=.98\textwidth]{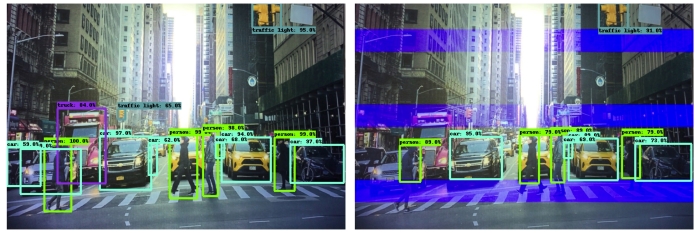}
		\label{fig:frcnn_cam_iphone}
		\caption{Apple iPhone 7}
	\end{subfigure}
	\begin{subfigure}[b]{0.49\textwidth}
		\centering
		\includegraphics[width=.98\textwidth]{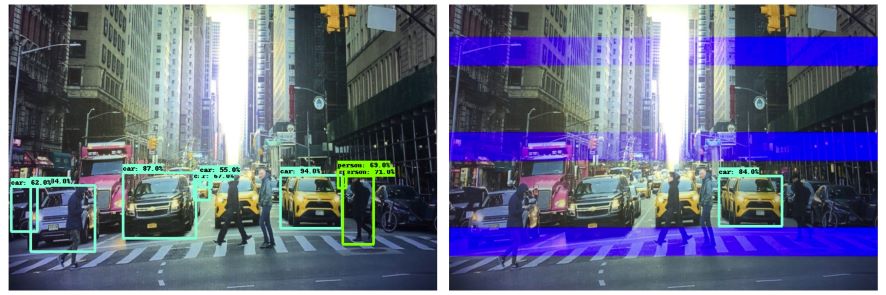}
		\label{fig:ssd_cam_iphone}
		\caption{Apple iPhone 7}
	\end{subfigure}
	\caption{Physical evaluation of the attack. These images were collected during a physical experiment where we printed images, placed them in front of the camera, and carried out the attack with the laser (see Figure~\ref{fig:experimental_setup}). The first column reports the object detection results on FRCNN, the second column on SSD.}
	\label{fig:results_cameras}
\end{figure*}

\end{appendices}

\end{document}